\documentclass{article}

\PassOptionsToPackage{numbers, compress}{natbib}

\usepackage[preprint]{neurips_2026}

\usepackage[utf8]{inputenc} 
\usepackage[T1]{fontenc}    
\usepackage{hyperref}       
\usepackage{url}            
\usepackage{booktabs}       
\usepackage{amsfonts}       
\usepackage{nicefrac}       
\usepackage{microtype}      
\usepackage{xcolor}         
\usepackage{enumitem}
\usepackage{amsmath}
\usepackage{graphicx}
\usepackage{caption}
\usepackage{subcaption}
\usepackage{wrapfig}
\usepackage{xcolor}
\usepackage[most]{tcolorbox}

\newtcolorbox{promptbox}[1]{
    colback=white,         
    colframe=black!75,     
    colbacktitle=black!75,  
    coltitle=white,         
    title={#1},            
    boxrule=1pt,            
    arc=2pt,                
    fonttitle=\rmfamily,    
    left=6pt, right=6pt, top=6pt, bottom=6pt
}

\title{Spatiotemporal Hidden-State Dynamics as a Signature of Internal Reasoning in Large Language Models}

\author{%
  Kotaro Furuya \quad Takahito Tanimura \\
  Research and Development Group, Hitachi, Ltd. \\
  1-280 Higashi-Koigakubo, Kokubunji-shi, Tokyo 185-8601, Japan \\
  \texttt{kotaro.furuya.ao@hitachi.com}
}

\newcommand{\stalt}{StALT}
\newcommand{\dtime}{\Delta_{\mathrm{time}}}
\newcommand{\dlayer}{\Delta_{\mathrm{layer}}}
\newcommand{\tdtime}{\tilde{\Delta}_{\mathrm{time}}}
\newcommand{\tdlayer}{\tilde{\Delta}_{\mathrm{layer}}}

\begin{document}

\maketitle

\begin{abstract}
Large reasoning models (LRMs) generate extended solutions, yet it remains unclear whether these traces reflect substantive internal computation or merely verbosity and overthinking.
Although recent hidden-state analyses suggest that internal representations carry correctness-related signals, their coarse aggregations may obscure the token and layer structure underlying reasoning computation.
We investigate hidden-state transitions across decoding steps and layers, and identify a distinct spatiotemporal pattern in LRMs: successful trajectories exhibit broad temporal dynamics with localized layer-wise concentration, while this structure is weaker in non-reasoning models and knowledge-heavy domains.
We formalize this characteristic as \emph{Spatiotemporal Amplitude of Latent Transition} (StALT), a training-free trajectory statistic that summarizes temporal changes between adjacent tokens weighted by within-token layer saliency. 
Across diverse models and benchmarks, StALT reliably separates correct from incorrect trajectories in reasoning-intensive regimes, providing a competitive label-free correctness signal alongside strong output-space and length-based baselines.
Intervention analyses further show that this spatiotemporal amplitude responds systematically to manipulations that increase or reduce the demand for internal reasoning, supporting its association with latent reasoning dynamics in LRMs.
These findings provide empirical evidence that LRMs exhibit measurable hidden-state dynamics and offer a practical probe for understanding internal computation beyond output-based evaluation.
\end{abstract}

\section{Introduction}

The paradigm of large language models (LLMs) has shifted from reactive text generation toward deliberate, multi-step reasoning.
Advances in test-time scaling, exemplified by chain-of-thought (CoT) \citep{wei2023chainofthoughtpromptingelicitsreasoning, snell2025scaling, muennighoff2025s1} and reinforcement learning methods designed to improve reasoning \citep{shao2024deepseekmath, Guo_2025}, have enabled LLMs to generate extended thinking trajectories before producing a final response. 
As large reasoning models (LRMs) push this capability further, evaluating their reasoning processes has become important for both reliable deployment \citep{huang2024trustllmtrustworthinesslargelanguage} and reward assignment during training \citep{lightman2023let}. 
However, current evaluation paradigms predominantly focus on outputs \cite{zhang2025sirenssongaiocean}, using answer confidence \citep{lin2022teachingmodelsexpressuncertainty, tian-etal-2023-just}, output logits \citep{agarwal2025the}, self-assessed certainty \citep{zhao2026learning, li2025confidenceneedfewshotrl}, external reward models \citep{lightman2023let}, or post-hoc multi-sample consistency \citep{xiong2024llmsexpressuncertaintyempirical}.
These tools are valuable for verification, but they leave the internal computation unobserved.
In particular, they do not address a more mechanistic question:
\textit{What happens inside the model while it is reasoning, and how does a successful trajectory diverge from an unsuccessful one?}

Addressing this question is vital, as longer reasoning traces do not necessarily indicate deliberate computation.
Models often suffer from ``overthinking'' \citep{zhang2025reasoningmodelsknowtheyre, eisenstadt2025overclockingllmreasoningmonitoring}, producing redundant tokens even after reaching the correct answer. 
Consequently, raw output length has been shown to be an unreliable proxy for reasoning quality \citep{wu2025lessunderstandingchainofthoughtlength}. 
A recent study therefore argues that reasoning-related computation may be more directly reflected in the hidden states than in surface trace length alone \citep{chen2026thinkdeepjustlong}.

Recognizing the value of internal signals, recent work has explored self-evaluation based on hidden states \citep{wang2025latent, liang2025cluenonparametricverificationexperience, ghasemabadi2026llmspredictfailuresselfawareness}. Most of these studies target relatively short generations or standard instruction-tuned models. \citet{vilas2025tracingtraceslatenttemporal} extend this line to reasoning models, showing that temporal latent-trajectory signals predict solution accuracy. 
However, their metrics aggregate over layers uniformly, without resolving the layer-local structure that recent analyses associate with active reasoning \citep{yang2026decoupling, wu2025knowledgereasoningcloselook}. 
If the trajectory is smoothed too aggressively over time or depth, transient dynamics such as backtracking and the localized computational hotspots that distinguish deliberation from passive recall can be obscured.
An operational probe therefore needs to summarize the trajectory without discarding its step-by-step structure.

Motivated by these findings, this work investigates the spatiotemporal dynamics of hidden states as empirical observables of internal computation, examining fluctuations across decoding steps and variation across layers.
Across open-weight models and domains, we identify a consistent empirical pattern. 
In reasoning models, internal deliberation manifests as broad temporal fluctuations and sharp layer-local hotspots that separate correct from incorrect trajectories.
This signature peaks on reasoning-intensive tasks and weakens on knowledge-retrieval tasks, suggesting deliberative multi-step computation rather than passive factual recall.

To operationalize this characteristic, we introduce \emph{Spatiotemporal Amplitude of Latent Transition} (\stalt{}), a scalable and training-free probe.
\stalt{} computes token-wise temporal amplitude and weights it by layer-local saliency, allowing the metric to retain localized structure while producing a single score.

Our contributions are as follows:
\begin{itemize}[leftmargin=*, itemsep=0.0em, topsep=0.0em, parsep=0pt, partopsep=0pt]
    \item We document a robust spatiotemporal pattern in hidden-state fluctuations that differentiates successful from unsuccessful trajectories in reasoning-intensive settings.
    \item We introduce \emph{Spatiotemporal Amplitude of Latent Transition} (\stalt{}), a simple training-free probe that quantifies reasoning-associated hidden-state dynamics by combining token-wise temporal amplitude with dynamic layer-local weighting.
    \item StALT exhibits especially higher values in reasoning models and on reasoning-intensive tasks, and responds systematically to interventions that modulate those values. These findings suggest that our probe can capture part of the internal computation underlying reasoning.
\end{itemize}

\section{Discovering Reasoning-Specific Hidden-State Dynamics}
\label{sec:observations}
As noted above, coarse trajectory summaries can smooth out the step-wise and layer-local structure that may characterize extended reasoning.
We therefore investigate whether hidden states evolve differently in reasoning models than in non-reasoning baselines, and whether that evolution distinguishes correct from incorrect trajectories at the level of individual layers and decoding steps.

\subsection{Spatiotemporal quantities}

To examine hidden-state dynamics at a spatiotemporal resolution, we define two elementary quantities that measure temporal change across decoding steps and spatial change across layers, respectively.

Consider a large language model with hidden layers $l \in \{0,\dots,L\}$ generating a response of length $T$.
We index generation steps by $t \in \{1,\dots,T\}$.
Let $h_t^l \in \mathbb{R}^D$ denote the hidden state at generation step $t$ and layer $l$, where $l=0$ is the embedding layer. 
Here, we exclude prompt-token positions from the trajectory.
We define two complementary forms of change:
\begin{align*}
    \dtime[t,l]
    &=
    \left\lVert h_t^l - h_{t-1}^l \right\rVert_2,
    \quad
    \text{for } t=2,\ldots,T,\, l=0,\ldots,L, \\
    \dlayer[t,l]
    &=
    \left\lVert h_t^l - h_t^{l-1} \right\rVert_2,
    \quad
    \text{for } t=1,\ldots,T,\, l=1,\ldots,L.
\end{align*}
Thus, $\dtime \in \mathbb{R}^{(T-1)\times(L+1)}, \dlayer \in \mathbb{R}^{T\times L}$.
Here, $\dtime$ measures how much the representation at a given layer changes between adjacent decoding steps, whereas $\dlayer$ measures how much the representation changes between adjacent layers at a given decoding step. 
These two quantities provide a token-by-layer grid of local variation that can reveal where and when the model's internal state undergoes the largest updates during generation.

To contrast reasoning and non-reasoning behavior, we examine Qwen3-0.6B \citep{yang2025qwen3technicalreport} as a representative reasoning model and Llama-3.1-8B-Instruct \citep{grattafiori2024llama3herdmodels} as a non-reasoning baseline. 
For datasets, we use integer problems from s1K-1.1 \citep{muennighoff2025s1}, which provide high-quality mathematical reasoning problems with verifiable correctness, and MMLU-Pro \citep{NEURIPS2024_ad236edc}, which spans reasoning-intensive and knowledge-oriented domains and allows us to test whether the observed dynamics are domain-dependent.

\subsection{Reasoning vs.\ non-reasoning dynamics}

\begin{figure*}[t]
\centering
\includegraphics[width=0.98\textwidth]{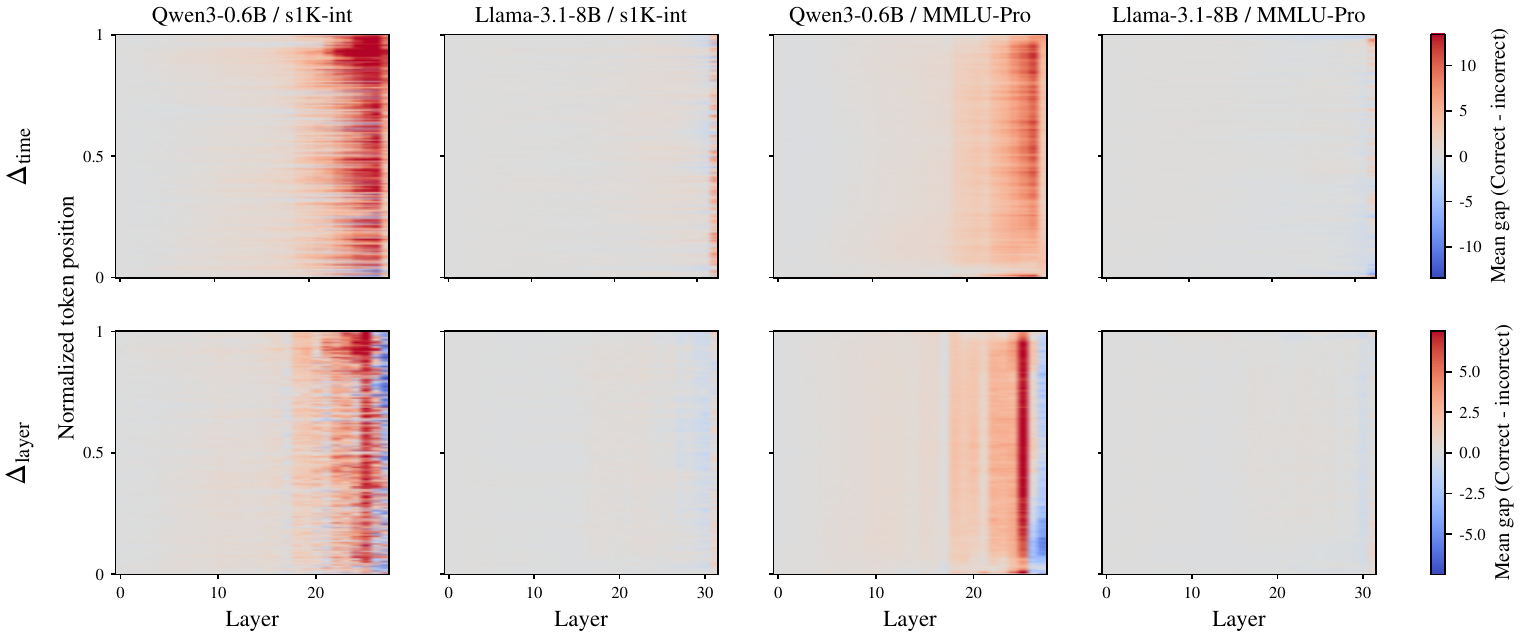}
\caption{Difference heatmaps between correct and incorrect hidden-state trajectories. Each cell shows the difference between the mean metric value over correct runs and the mean over incorrect runs, at a given layer and relative token position. The top row shows time-wise changes ($\dtime$) and the bottom row shows layer-wise changes ($\dlayer$).}
\label{fig:s1k_numeric_obs}
\vspace{-1.\baselineskip}
\end{figure*}

Figure~\ref{fig:s1k_numeric_obs} visualizes the differences in hidden-state dynamics between correct and incorrect generations for integer problems from s1K-1.1 and MMLU-Pro, comparing the reasoning model with the non-reasoning baseline.
Because reasoning traces vary in length, we normalize each trajectory to relative token positions via linear interpolation, allowing correct and incorrect runs to be compared on a common axis.
The heatmap shows the difference between the mean value over correct runs and the mean value over incorrect runs at each layer and normalized token position.

For the reasoning model, the temporal change $\dtime$ shows a clear separation between correct and incorrect trajectories, concentrated near the end of generation and spread across a broad range of higher layers.
The layer-wise change $\dlayer$ exhibits a similar late-stage contrast, but the difference is more sharply localized in specific layers rather than distributed broadly.
This asymmetry suggests that the two quantities capture complementary aspects of the dynamics. 
Temporal variation distinguishes correct from incorrect trajectories across a broad range of layers, whereas layer-wise variation does so at a few specific depths.
Overall, both measures tend to be larger along correct trajectories than along incorrect ones.

In contrast, the non-reasoning baseline exhibits more uniform dynamics.
Neither $\dtime$ nor $\dlayer$ shows the localized hotspots observed in the reasoning model, and the gap between correct and incorrect trajectories remains small across layers and token positions.

\subsection{Domain dependence of the dynamics}

We next ask whether the observed dynamics reflect a generic property of model generation or whether they are selective to domains that require multi-step reasoning.
MMLU-Pro provides a useful natural intervention, as its subjects range from reasoning-intensive STEM domains to more knowledge-oriented Non-STEM domains while sharing a common multiple-choice format.
We therefore group its subjects into three coarse categories: STEM (biology, chemistry, physics, math, engineering, and computer science), Non-STEM (business, economics, health, history, law, philosophy, and psychology), and Other.
For each group, we compute the correctness-associated dynamical gap as the Hedges' $g$ effect size \citep{hedges1981distribution}, enabling fair comparisons across domains and models.

\begin{figure*}[t]
\centering
\includegraphics[width=0.9\textwidth]{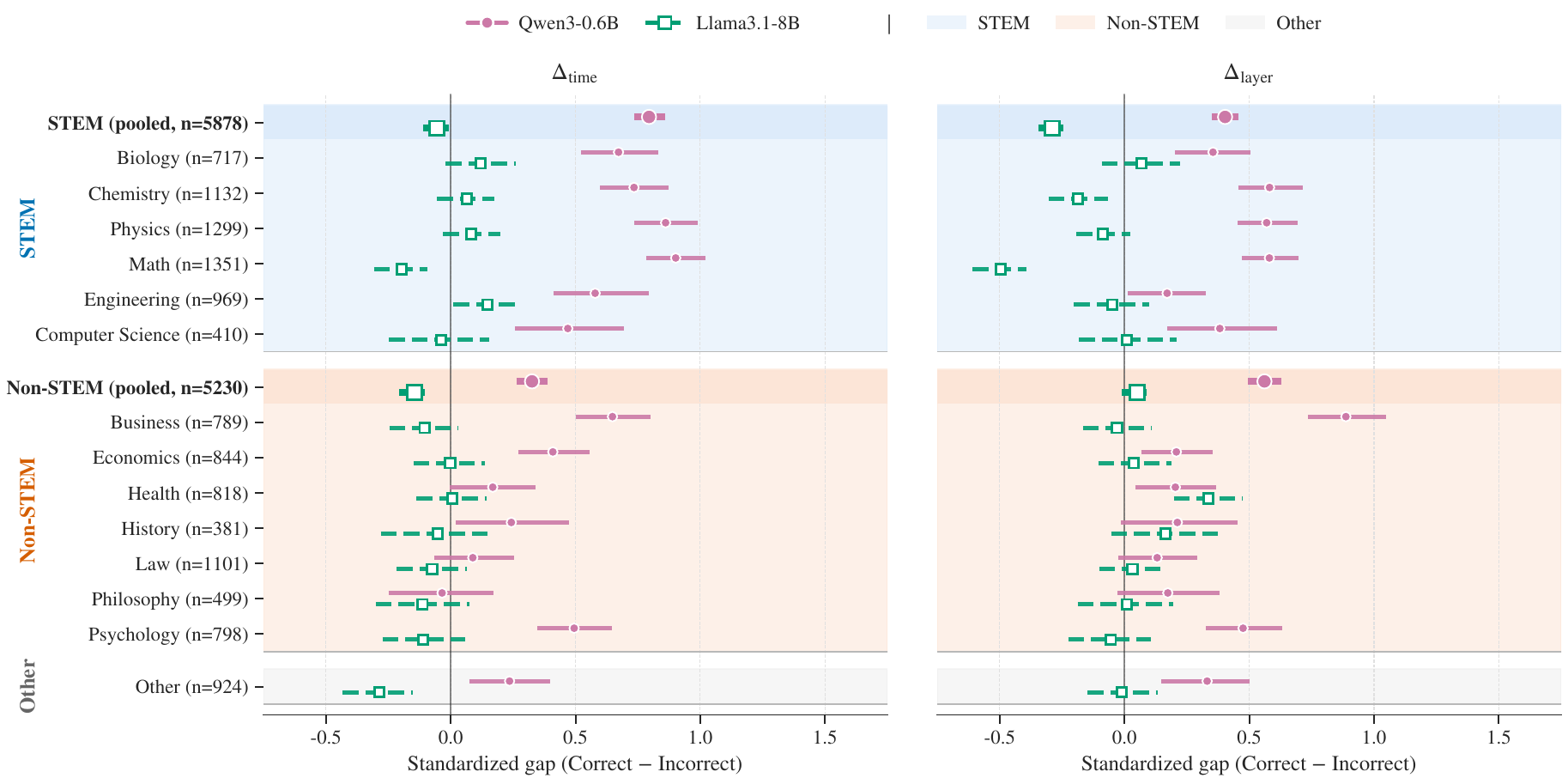}
\caption{Standardized gap in hidden-state dynamics between correct and incorrect trajectories on MMLU-Pro. Horizontal error bars denote 95\% bootstrap confidence intervals computed with 4,000 resamples.}
\label{fig:mmlupro_qwen_obs}
\vspace{-1.5\baselineskip}
\end{figure*}

Figure~\ref{fig:mmlupro_qwen_obs} shows the standardized gap in hidden-state dynamics between correct and incorrect trajectories at both the group and subject levels.
In the reasoning model, the gap varies substantially across domains, being largest in STEM and much smaller in Non-STEM and Other.
At the subject level, reasoning-intensive fields such as mathematics, physics, chemistry, and biology exhibit the strongest separation, whereas more knowledge-oriented subjects such as history and law show much weaker effects.

By contrast, the non-reasoning baseline exhibits a uniformly narrow gap across all three groups, with the overall magnitude of the dynamical difference remaining small regardless of subject. 
This contrast suggests that the spatiotemporal pattern is not a uniform correlate of answer correctness, but is selectively amplified in settings that require internally sustained reasoning.

\subsection{Observation summary and hypothesis}

Three patterns emerge from the observations above. 
First, in the reasoning model, correct trajectories exhibit larger spatiotemporal hidden-state dynamics than incorrect ones, and this contrast is much weaker in the non-reasoning baseline.
Second, the two quantities play asymmetric roles. $\dtime$ separates correct from incorrect trajectories across a broad range of layers, whereas $\dlayer$ separates them at a few concentrated depths.
Third, this dynamical gap is domain-dependent, appearing most strongly on reasoning-intensive subjects and weakening on knowledge-oriented ones.

These concurrent patterns motivate our core empirical hypothesis: \textit{latent reasoning is associated with structured spatiotemporal variation in hidden-state trajectories, characterized by broad temporal fluctuation modulated by localized layer-wise concentration.}
The asymmetry between $\dtime$ and $\dlayer$ also suggests a design principle for summarizing this signature. 
Temporal change appears to be the primary signal, while layer-wise variation indicates where that signal is most concentrated.
In the next section, we formalize this principle as a scalar probe of the amplitude of latent transitions across time and layers.

\section{Quantifying the Signature with \stalt{}}
\label{sec:metric}

\begin{figure}[t]
    \centering
    \includegraphics[width=0.8\linewidth]{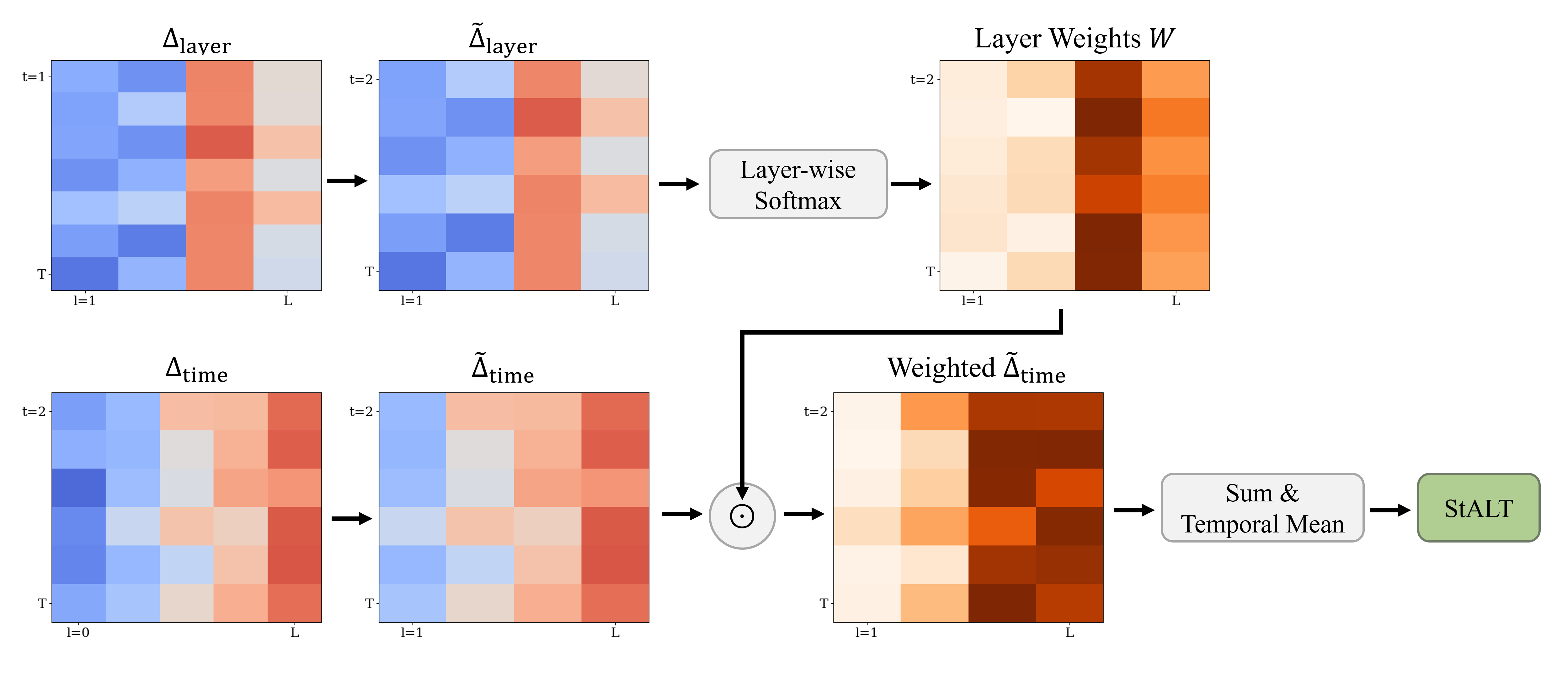}
    \caption{Conceptual overview of \stalt. Temporal changes ($\tdtime$) are weighted by layer-local saliency ($\tdlayer$) to emphasize where and when internal computation is most active.}
    \label{fig:lwtd}
    \vspace{-1.\baselineskip}
\end{figure}

Following the design principle identified in Section~\ref{sec:observations}, we now formalize a metric that treats temporal change as the primary signal and uses layer-wise variation as a selector over layers. 
We introduce \emph{Spatiotemporal Amplitude of Latent Transition} (\stalt{}), a training-free probe of reasoning-associated hidden-state dynamics.
A summary statistic for the observed dynamics should satisfy three properties: (1) temporal sensitivity, preserving the token-by-token evolution of hidden states rather than compressing the trajectory to an endpoint; (2) layer selectivity, focusing on the depths where computation is most concentrated rather than averaging uniformly across layers; and (3) unsupervised applicability, requiring no labels, probes, or task-specific training to compute the statistic.

\subsection{Definition}
Figure~\ref{fig:lwtd} provides an overview of the construction.
We begin by aligning the tensor dimensions by removing the embedding column from $\dtime$ and the initial time row from $\dlayer$.
From $\dtime$, we drop the embedding-layer column ($l=0$), whose temporal changes reflect token-identity shifts rather than computational updates.
From $\dlayer$, we drop the first-token row ($t=1$), which has no corresponding entry in $\dtime$.
The resulting aligned tensors $\tdtime,\tdlayer\in\mathbb{R}^{(T-1)\times L}$ share the same indices over steps $t=2,\dots,T$ and layers $l=1,\dots,L$.
For each time step $t$, we convert the aligned layer-local changes into layer weights:
\begin{equation*}
    W[t,l]
    =
    \frac{
        \exp(\tdlayer[t,l]/\tau)
    }{
        \sum_{k=1}^{L}
        \exp(\tdlayer[t,k]/\tau)
    }.
\end{equation*}
This assigns higher weight to layers undergoing the largest within-token representational shifts, implementing the layer-selection role of $\dlayer$ identified in Section~\ref{sec:observations}. 
We then define \stalt{} as the time-averaged, layer-weighted temporal amplitude:
\begin{equation*}
    \mathrm{StALT}
    =
    \frac{1}{T-1}
    \sum_{t=2}^{T}
    \sum_{l=1}^{L}
    W[t,l]\cdot \tdtime[t,l].
\end{equation*}
Thus, \stalt{} assigns a single scalar score to each generated trajectory.

\subsection{Rationale for the Aggregation Order}

The asymmetry between temporal and layer-wise roles is a deliberate design choice motivated by the observations in Section~\ref{sec:observations}, where $\dtime$ carried the broadly distributed correctness signal while $\dlayer$ localized it to specific depths.
We also consider the reverse aggregation order in Appendix~\ref{app:aggregation_order}, where time-wise saliency is used to weight layer-wise amplitude.
This variant is generally weaker, supporting the interpretation that the observed signature is better characterized as temporal movement with layer-local concentration, rather than layer-wise movement selected over time.

\subsection{Interpretation}

Higher \stalt{} values indicate larger temporal hidden-state amplitude in layers with strong layer-local transformations. 
We interpret \stalt{} as a descriptive trajectory-level statistic and expect it to be most informative in regimes where successful generation requires extended internal computation.
Unless otherwise stated, we set the temperature to $\tau=1$; Appendix~\ref{app:metric_ablations} reports a sweep over $\tau$ and discusses its effect on layer selection.

\section{Large-Scale Evaluation of \stalt{}}
\label{sec:large_scale}

We begin our evaluation by asking whether our hypothesis with  \stalt{} scales from the discovery setting to a broad range of models and tasks. 
The central prediction is that the spatiotemporal signature should emerge in reasoning models and reasoning-intensive tasks, where successful generation depends on internal computation.
Our evaluation has two parts. 
First, we measure the standardized gap between correct and incorrect trajectories in \stalt{} to identify the model and task regimes in which the hidden-state signature is most strongly expressed.
Second, we evaluate whether the same statistic is also useful as a label-free correctness predictor against diverse baselines.

\subsection{Evaluation Setup}

\paragraph{Datasets.}
We select datasets to probe different aspects of reasoning.
To measure mathematical reasoning across varying difficulty levels, we use GSM8K \citep{cobbe2021gsm8k} and MATH-500 \citep{math500}.
To assess reasoning capabilities in a highly complex domain outside of mathematics, we include GPQA-Diamond \citep{rein2024gpqa}. 
We also evaluate on MMLU-Pro \citep{NEURIPS2024_ad236edc}, separating the STEM and Non-STEM subsets to contrast domain-specific performance between reasoning-intensive problems and knowledge-retrieval tasks.
\vspace{-0.5\baselineskip}

\paragraph{Models.}
We compare five reasoning models, Qwen3-1.7B, Qwen3-4B, Qwen3-8B, SmolLM3-3B, and gpt-oss-20b \citep{yang2025qwen3technicalreport,bakouch2025smollm3, openai2025gptoss120bgptoss20bmodel}, against non-reasoning models. 
For contrast, we also include Qwen2.5-Math-7B \citep{yang2024qwen25mathtechnicalreportmathematical} and Llama-3.1-8B-Instruct \citep{grattafiori2024llama3herdmodels} as non-reasoning baselines.
For the reasoning models, generation is performed with thinking mode; specifically, for gpt-oss-20b, we set the thinking effort to ``medium.'' 
Across all datasets, models are evaluated in a zero-shot setting with CoT prompts \citep{wei2023chainofthoughtpromptingelicitsreasoning, 10.5555/3600270.3601883}. 
Implementation details are provided in Appendix~\ref{app:experimental_setup}.
\vspace{-0.5\baselineskip}

\paragraph{Baselines.}
We compare \stalt{} against four families of label-free predictors, (1) generated token count, motivated by prior work suggesting that excessively long reasoning traces tend to be associated with incorrect answers \citep{wu2025lessunderstandingchainofthoughtlength, chen2026thinkdeepjustlong}; (2) output-space confidence proxies such as maximum token probability, perplexity, and entropy; (3) embedding-space verification metrics, specifically CoE-R and CoE-C \citep{wang2025latent}; and (4) the unweighted means of $\dtime$ and $\dlayer$, to ablate the necessity of our layer-weighted temporal integration.
\vspace{-0.5\baselineskip}

\paragraph{Evaluation.}
Following the evaluation protocol established in Wang et al. \citep{wang2025latent}, each metric is evaluated as an unsupervised correctness predictor.
We report AUROC as the primary metric in the main text.
FPR95 and AUPR are reported in Appendix~\ref{app:full_results}.
Since generated token count, perplexity, and entropy are negatively correlated with correctness, we negate these scores to compute the evaluation metrics.
Results are averaged over five independent runs to ensure robustness.

\subsection{Results}

\begin{figure}[t]
\centering
\includegraphics[width=0.5\linewidth]{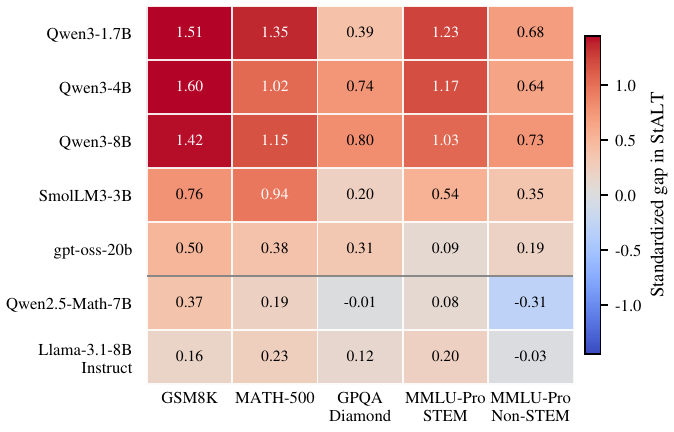}
\caption{Standardized gap between correct and incorrect trajectories, measured by StALT, across model and task conditions.}
\label{fig:stalt_gap_heatmap}
\vspace{-1.\baselineskip}
\end{figure}

Figure~\ref{fig:stalt_gap_heatmap} examines whether \stalt{} differs between successful and unsuccessful trajectories.
Each cell reports the standardized gap (Hedges' $g$) in \stalt{} between correct and incorrect trajectories within each model and task.

The largest positive gaps appear in reasoning models on reasoning-intensive tasks, especially GSM8K and MATH-500.
MMLU-Pro STEM also exhibits strong positive gaps, indicating that the signature is not limited to mathematical reasoning but extends to other reasoning-intensive domains.
GPQA-Diamond, which requires the application of domain knowledge, provides a boundary case among the datasets, yielding smaller gaps relative to other reasoning-intensive datasets and indicating the gradient of the dynamics across reasoning and knowledge regimes.
On the model side, the effect is strongest in the Qwen3 family, where the signature was first observed in Section~\ref{sec:observations}.
Importantly, however, the same tendency is also visible in other LRMs, including SmolLM3-3B and gpt-oss-20b.
This suggests that the observed separation is not specific to a single model family, but can reflect a broader property of reasoning models.

The gap is weaker in non-LRM baselines and on MMLU-Pro Non-STEM.
Qwen2.5-Math-7B provides a useful boundary case. 
Although it is specialized for mathematical problem solving, it does not exhibit the same consistent positive gap as reasoning-oriented models.
This suggests that task specialization alone is insufficient to produce the dynamical signature.
Overall, these results suggest that StALT appears to be larger in reasoning regimes that place greater demands on deliberative internal computation.

\begin{figure*}[t]
\centering
\begin{subfigure}[t]{0.32\textwidth}
    \centering
    \includegraphics[width=\linewidth]{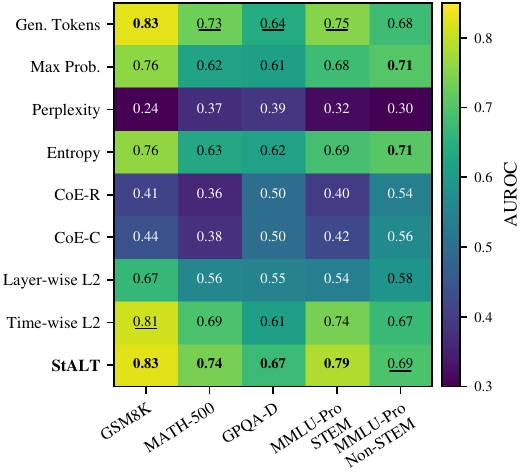}
    \caption{LRMs: Qwen3 family}
    \label{fig:auc_qwen3}
\end{subfigure}
\hfill
\begin{subfigure}[t]{0.32\textwidth}
    \centering
    \includegraphics[width=\linewidth]{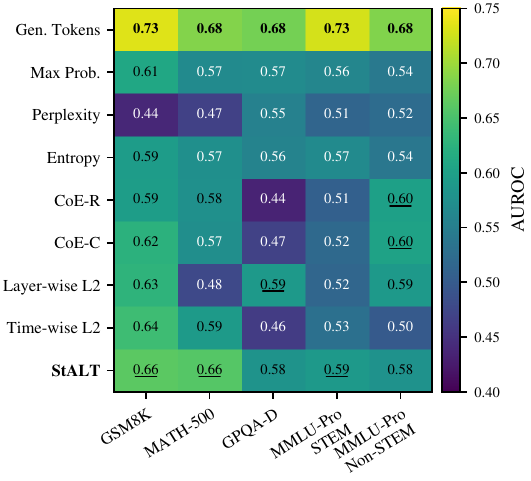}
    \caption{LRMs: other families}
    \label{fig:auc_other_lrm}
\end{subfigure}
\hfill
\begin{subfigure}[t]{0.32\textwidth}
    \centering
    \includegraphics[width=\linewidth]{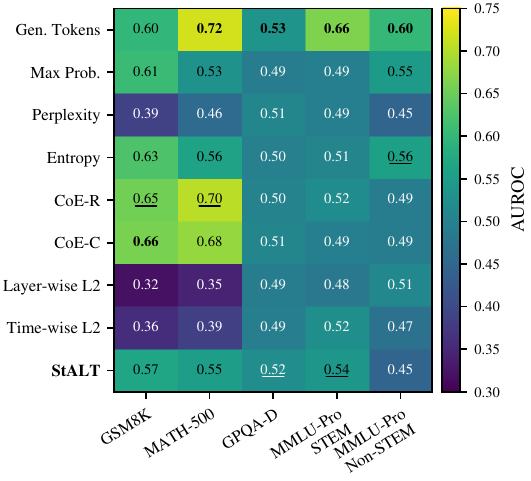}
    \caption{Non-LRM baselines}
    \label{fig:auc_non_lrm}
\end{subfigure}
\caption{Mean AUROC of \stalt{} and representative baselines averaged within model groups. Group-specific color scales are used to emphasize within-group method comparisons. Qwen3 LRMs include Qwen3-1.7B, Qwen3-4B, and Qwen3-8B; other LRMs include SmolLM3-3B and gpt-oss-20b; non-LRM baselines include Qwen2.5-Math-7B and Llama-3.1-8B-Instruct.}
\label{fig:aggregated_baseline_auc}
\vspace{-1.5\baselineskip}
\end{figure*}

Figure~\ref{fig:aggregated_baseline_auc} next asks whether the same statistic is also useful for label-free correctness prediction.
To make the comparison interpretable, we aggregate results into three model groups: Qwen3 LRMs, other LRMs, and non-LRM baselines.
This grouping separates the family in which the signature was first observed from other reasoning models, while avoiding a single strong family from dominating the overall LRM trend.

\begin{wrapfigure}{r}{0.4\columnwidth}
    \vspace{-1.4\baselineskip}
    \centering
    \includegraphics[width=0.4\columnwidth]{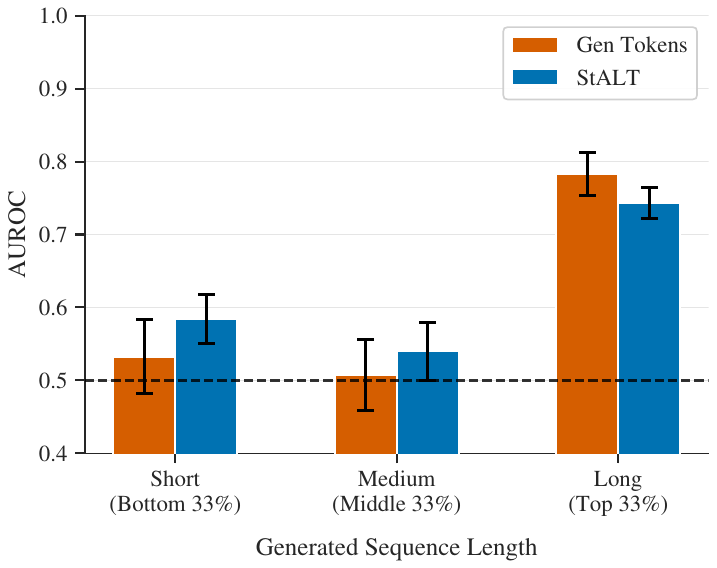}
    \caption{Length-stratified AUROC on MATH-500 for Qwen3-4B. Error bars denote one standard deviation across runs.}
    \vspace{-1.6\baselineskip}
    \label{fig:length_stratified_qwen34b_math500}
\end{wrapfigure}

In the LRM groups, \stalt{} is competitive with strong label-free predictors, particularly on reasoning-intensive datasets such as GSM8K, MATH-500, and MMLU-Pro STEM.
The effect is clearest in the Qwen3 group, but the same tendency remains in other LRMs, although with smaller margins.
By contrast, the non-LRM baselines show weaker and less consistent gains from \stalt{}.

Generated-token count remains a competitive baseline.
This raises a potential confound: reasoning models tend to produce shorter traces on problems they solve correctly, suggesting that \stalt{} may simply be capturing response length.  
However, Figure~\ref{fig:length_stratified_qwen34b_math500} shows that \stalt{} remains predictive even within fixed generated length.
This weakens the interpretation that the metric is merely rediscovering response length.
Rather, \stalt{} appears to capture a dynamical property of the trajectory that persists within coarse length-controlled subsets.

These results empirically extend our observations on hidden-state dynamics from the discovery setting to reasoning models and reasoning-intensive tasks. 
The signal is strongest in regimes where successful generation is likely to require extended internal computation, and weakens in settings where correctness is more plausibly supported by recall, domain knowledge, or task-specific specialization.
This suggests that \stalt{} is not a generic correctness proxy, but a label-free signal associated with internal reasoning computation.

\section{Intervention Analyses of Internal Reasoning}
\label{sec:perturbation}

Large-scale evaluation showed that the hidden-state signature is strongest in reasoning regimes, but cross-model comparisons confound objective, scale, and architecture.
To more directly probe the role of latent dynamics, we conduct intervention analyses using Qwen3-4B as a fixed backbone with integer problems from s1K-1.1.
The section is organized around a directional prediction: reasoning-amplifying interventions should raise \stalt{}, whereas reasoning-reducing interventions should lower it.
Unless otherwise noted, we report \stalt{}, accuracy, and mean generated length over five runs.

\subsection{Reasoning-Amplifying Interventions}

\begin{wrapfigure}{r}{0.45\columnwidth}
    \vspace{-3.\baselineskip}
    \centering
    \includegraphics[width=0.45\textwidth]{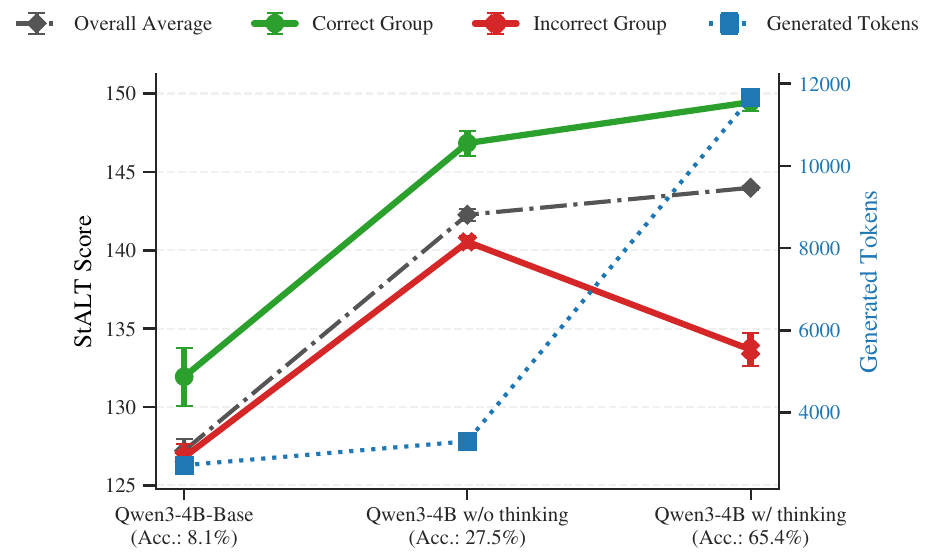}
    \caption{Within-family comparison on s1K-1.1 for Qwen3-4B across base, no-thinking, and thinking-mode settings.}
    \vspace{-1.4\baselineskip}
    \label{fig:thinking_mode}
\end{wrapfigure}

We first consider interventions that should amplify internal dynamics, either by tuning the model for reasoning or by activating thinking mode at inference time.
We compare Qwen3-4B-Base, Qwen3-4B with thinking mode disabled, and Qwen3-4B with thinking mode enabled. 
If our probe captures latent reasoning, it should increase across these conditions.

Figure~\ref{fig:thinking_mode} shows the resulting \stalt{} scores together with accuracy and mean generated length on the integer subset of s1K-1.1.
\stalt{} increases monotonically from the base model to the reasoning-tuned model without thinking, and increases further when thinking mode is enabled.
Notably, the shift from the base model to the no-thinking setting yields only a modest increase in generated length, while both \stalt{} and accuracy rise clearly.
This suggests that reasoning-oriented tuning itself increases hidden-state variation even before explicit extended thinking is activated.
Enabling thinking mode then produces a much larger increase in output length together with a further strengthening of the hidden-state signature and task performance.
Consequently, these results indicate that interventions promoting internal reasoning simultaneously amplify the hidden-state signature, supporting the interpretation of \stalt{} as a marker of internal computation.

\subsection{Reasoning-Reducing Interventions}
We next examine interventions that reduce internal reasoning demand, either by externalizing reasoning through in-context CoT exemplars or by supervised fine-tuning.

\subsubsection{In-Context CoT Scaffolding}

\begin{wrapfigure}{r}{0.45\columnwidth}
    \vspace{-3.\baselineskip}
    \centering
    \includegraphics[width=0.45\textwidth]{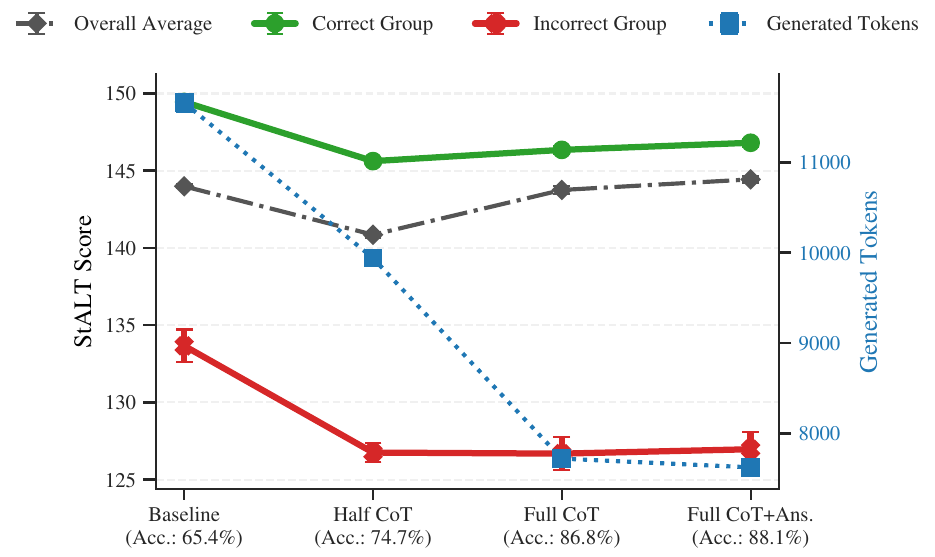}
    \caption{Effect of in-context CoT scaffolding on hidden-state dynamics for Qwen3-4B.}
    \vspace{-1.4\baselineskip}
    \label{fig:cot}
\end{wrapfigure}

We first consider prompt-space interventions. 
Instead of encouraging the model to reason more internally, we provide part of the solution process externally.
We use the DeepSeek-R1 reasoning traces and answers included in s1K-1.1 as externally supplied solution traces and compare four prompting conditions: a baseline prompt without reasoning exemplars, half CoT (the first half of the tokens in the trace), full CoT, and full CoT plus the answer.
If \stalt{} tracks internally executed reasoning, supplying intermediate steps in the prompt should reduce the need for internal computation.

Figure~\ref{fig:cot} shows the results for Qwen3-4B.
As more of the solution is provided externally, accuracy increases monotonically and generated length decreases sharply.
Mean \stalt{} declines under partial scaffolding but recovers close to the baseline under full scaffolding.
This apparent recovery is driven by a change in group composition: both the correct and incorrect groups individually show declining \stalt{} as more reasoning is externalized, while the improvement in accuracy increases the share of correct cases, which have higher \stalt{}.
Therefore, external scaffolding substitutes for internal computation rather than increasing it.

\subsubsection{Supervised Fine-Tuning}
\begin{wrapfigure}{r}{0.45\columnwidth}
    \vspace{-3.\baselineskip}
    \centering
    \includegraphics[width=0.45\textwidth]{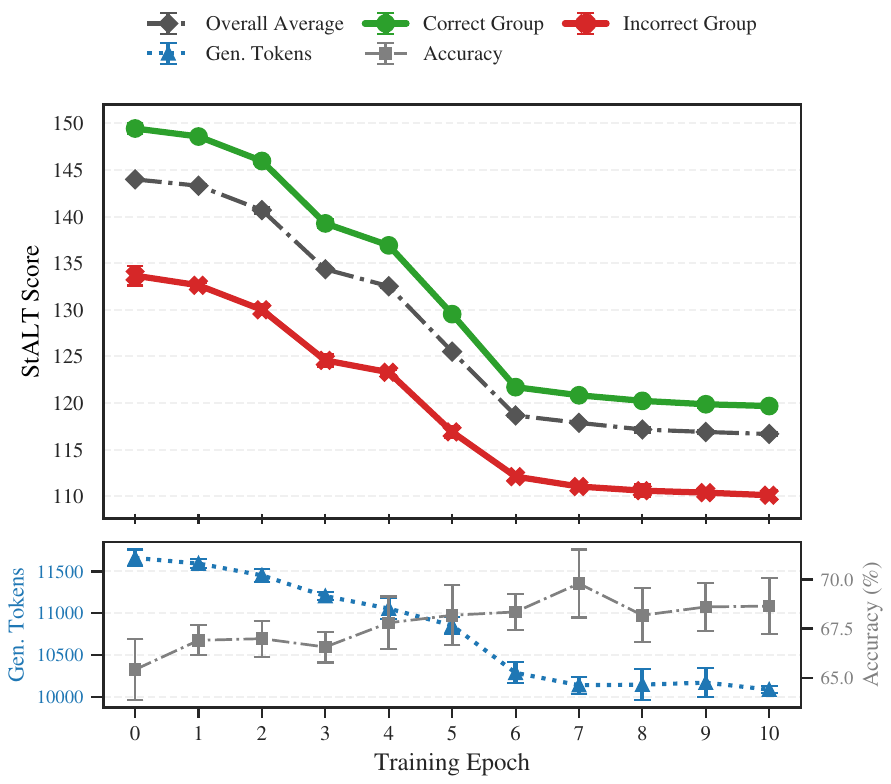}
    \caption{Effect of supervised fine-tuning on hidden-state dynamics for s1K-1.1.}
    \vspace{-1.4\baselineskip}
    \label{fig:sft}
\end{wrapfigure}

Supervised fine-tuning provides a second reasoning-reducing intervention, but in parameter space.
We train Qwen3-4B on the DeepSeek-R1 reasoning traces and responses for the integer problems in s1K-1.1, and evaluate the resulting checkpoints on the same set of problems. 
This allows us to examine how parameterizing the solution traces changes the hidden-state signature in the regime where those traces are directly available during fine-tuning.
We expect that fine-tuning on the specific solutions should reduce the hidden-state signature even when in-domain performance improves.

Figure~\ref{fig:sft} tracks \stalt{} and accuracy over checkpoints for Qwen3-4B. 
Accuracy improves slightly while \stalt{} declines steadily. 
This divergence between performance and hidden-state dynamics is consistent with the idea that fine-tuning makes the model more reliant on memorized patterns and less reliant on internal updates. 
Mean generated token length also shows an overall decline over training, but more gradually than \stalt{}, indicating that the weakening of the internal signature is not simply a length effect.
The results are consistent with prior findings that supervised fine-tuning can boost recall while dampening deliberative reasoning paths \citep{wu2025knowledgereasoningcloselook}.

These intervention analyses support a consistent directional pattern: interventions that encourage internal reasoning amplify the signature, whereas interventions that externalize reasoning or shift computation into learned patterns attenuate it.

\section{Discussion and Limitations}

Our results characterize a distinctive spatiotemporal pattern in the hidden-state trajectories of LRMs during successful reasoning.
The pattern is most pronounced in reasoning models and reasoning-intensive tasks, while it weakens in knowledge-heavy domains and non-reasoning models.
These results suggest that hidden states are not merely predictive of correctness, but reflect a structured form of latent computation that differs across model and task regimes.

\stalt{} provides a compact way to quantify this structure.
By aggregating how internal representations evolve across layers and decoding steps, \stalt{} captures the temporal organization of latent computation.
This view connects naturally to interpretations of CoT reasoning as an iterative refinement process \citep{huang2025transformers}, in which intermediate tokens progressively reshape the model's internal state.
From this perspective, the observed trajectory pattern offers a useful internal lens for self-evaluation, label-free reward for reinforcement learning, and mechanistic analysis of reasoning dynamics.

Several practical limitations remain.
The method requires access to full hidden states, restricting its applicability to open-weight models.
Moreover, although our intervention analyses support a systematic association between the observed trajectory patterns and reasoning demand, they do not isolate a single causal mechanism.
Our findings therefore identify and quantify a distinctive hidden-state signature of LRM reasoning, while leaving a complete mechanistic account of how this signature arises to future work.

\section{Conclusion}

We identified spatiotemporal hidden-state dynamics associated with successful reasoning trajectories in LRMs.
This signature generalizes beyond the initial Qwen3 observation to other LRMs, is strongest on reasoning-intensive tasks, and weakens in non-reasoning or knowledge-heavy regimes.
We formalized the pattern as \emph{Spatiotemporal Amplitude of Latent Transition} (\stalt{}), a training-free statistic for summarizing trajectory-level hidden-state dynamics.
Beyond label-free verification, our framework provides a mechanistic lens for examining the latent deliberation phase of LRMs, advancing the understanding of their hidden reasoning.

\bibliographystyle{unsrtnat}
\bibliography{hidden_state_verifi_shorten}

\newpage

\appendix

\section{Supplementary Results}
\label{app:intervention_supplement}

\subsection{Full Results}
\label{app:full_results}

Table~\ref{tab:comprehensive_results} reports the complete AUROC, FPR95, and AUPR results for all methods, models, and datasets discussed in Section~\ref{sec:large_scale}.

\begin{table*}[t]
\centering
\resizebox{\textwidth}{!}{
\begin{tabular}{l|ccccc|cc}
\toprule
\textbf{Method / Model} & \textbf{Qwen3-1.7B} & \textbf{Qwen3-4B} & \textbf{Qwen3-8B} & \textbf{SmolLM3-3B} & \textbf{gpt-oss-20b} & \textbf{Qwen2.5-Math-7B} & \textbf{Llama-3.1-8B-Instruct} \\ \midrule
\multicolumn{8}{c}{\textbf{MATH-500 (AUROC $\uparrow$ / FPR95 $\downarrow$ / AUPR $\uparrow$)}} \\ \midrule
Gen. Tokens & $\mathbf{0.78} / \underline{0.60} / \underline{0.91}$ & $\mathbf{0.71} / \mathbf{0.64} / \mathbf{0.92}$ & $0.70 / \mathbf{0.65} / \underline{0.92}$ & $\mathbf{0.74} / \mathbf{0.63} / \underline{0.91}$ & $\mathbf{0.63} / \mathbf{0.80} / \mathbf{0.87}$ & $0.65 / \mathbf{0.57} / 0.47$ & $\mathbf{0.79} / \mathbf{0.61} / \mathbf{0.69}$ \\
Max Prob. & $0.64 / 0.90 / 0.86$ & $0.63 / 0.92 / 0.90$ & $0.59 / 0.93 / 0.89$ & $0.64 / 0.89 / 0.88$ & $0.50 / 0.93 / 0.81$ & $0.52 / 0.83 / 0.41$ & $0.54 / 0.88 / 0.41$ \\
Perplexity & $0.35 / 0.98 / 0.74$ & $0.36 / 0.98 / 0.81$ & $0.40 / 0.97 / 0.82$ & $0.35 / 0.98 / 0.76$ & $0.59 / 0.95 / 0.85$ & $0.47 / 0.78 / 0.37$ & $0.44 / 0.87 / 0.36$ \\
Entropy & $0.65 / 0.91 / 0.87$ & $0.64 / 0.92 / 0.90$ & $0.60 / 0.93 / 0.89$ & $0.66 / 0.88 / 0.89$ & $0.49 / 0.96 / 0.81$ & $0.55 / 0.79 / 0.43$ & $0.57 / 0.86 / 0.43$ \\
CoE-R & $0.37 / 0.94 / 0.73$ & $0.42 / 0.93 / 0.81$ & $0.29 / 0.99 / 0.76$ & $\underline{0.72} / 0.82 / \mathbf{0.92}$ & $0.43 / 0.95 / 0.78$ & $\mathbf{0.73} / \underline{0.71} / \underline{0.60}$ & $\underline{0.68} / \underline{0.77} / \underline{0.56}$ \\
CoE-C & $0.40 / 0.94 / 0.75$ & $0.44 / 0.94 / 0.82$ & $0.29 / 0.99 / 0.76$ & $0.71 / 0.70 / \underline{0.91}$ & $0.43 / 0.96 / 0.78$ & $\underline{0.72} / 0.77 / \mathbf{0.61}$ & $0.64 / 0.84 / 0.50$ \\
Layer-wise $L_2$ & $0.59 / 0.89 / 0.84$ & $0.56 / 0.94 / 0.87$ & $0.51 / 0.97 / 0.86$ & $0.41 / 0.95 / 0.79$ & $0.54 / 0.97 / 0.83$ & $0.34 / 0.94 / 0.33$ & $0.35 / 0.99 / 0.34$ \\
Time-wise $L_2$ & $0.72 / 0.74 / 0.88$ & $\underline{0.69} / 0.80 / \mathbf{0.92}$ & $0.67 / 0.83 / 0.91$ & $0.63 / 0.92 / 0.87$ & $0.55 / 0.95 / 0.84$ & $0.37 / 0.93 / 0.34$ & $0.40 / 0.96 / 0.36$ \\
\textbf{StALT (Ours)} & $\mathbf{0.78} / \mathbf{0.58} / \mathbf{0.92}$ & $\underline{0.69} / \underline{0.66} / \mathbf{0.92}$ & $\underline{0.74} / \underline{0.69} / \mathbf{0.94}$ & $\underline{0.72} / \underline{0.67} / \underline{0.91}$ & $\underline{0.60} / \underline{0.86} / 0.85$ & $0.54 / 0.84 / 0.42$ & $0.56 / 0.87 / 0.45$ \\
\midrule
\multicolumn{8}{c}{\textbf{GSM8K (AUROC $\uparrow$ / FPR95 $\downarrow$ / AUPR $\uparrow$)}} \\ \midrule
Gen. Tokens & $\mathbf{0.82} / \mathbf{0.58} / \mathbf{0.97}$ & $\underline{0.83} / \mathbf{0.55} / \mathbf{0.98}$ & $\mathbf{0.83} / \mathbf{0.57} / \mathbf{0.99}$ & $\mathbf{0.77} / \mathbf{0.65} / \mathbf{0.96}$ & $\mathbf{0.69} / \mathbf{0.78} / \mathbf{0.94}$ & $0.53 / 0.74 / 0.52$ & $\mathbf{0.67} / \underline{0.80} / \mathbf{0.82}$ \\
Max Prob. & $0.76 / 0.80 / 0.95$ & $0.77 / 0.68 / \underline{0.97}$ & $0.74 / 0.77 / 0.97$ & $0.70 / \underline{0.76} / 0.94$ & $0.52 / 0.95 / 0.90$ & $0.59 / 0.71 / 0.57$ & $0.62 / \underline{0.80} / 0.78$ \\
Perplexity & $0.24 / 0.98 / 0.81$ & $0.22 / 0.97 / 0.87$ & $0.26 / 0.97 / 0.89$ & $0.30 / 0.98 / 0.82$ & $0.58 / 0.91 / 0.92$ & $0.40 / 0.85 / 0.45$ & $0.38 / 0.94 / 0.65$ \\
Entropy & $0.76 / 0.80 / 0.95$ & $0.78 / 0.71 / \underline{0.97}$ & $0.74 / 0.80 / 0.97$ & $0.71 / \underline{0.76} / 0.94$ & $0.48 / 0.97 / 0.89$ & $0.62 / \mathbf{0.69} / 0.59$ & $\underline{0.63} / \mathbf{0.78} / 0.78$ \\
CoE-R & $0.41 / 0.94 / 0.85$ & $0.52 / 0.93 / 0.93$ & $0.30 / 0.96 / 0.90$ & $0.67 / 0.85 / 0.94$ & $0.52 / 0.95 / 0.89$ & $\mathbf{0.75} / 0.73 / \mathbf{0.74}$ & $0.56 / 0.89 / 0.77$ \\
CoE-C & $0.46 / 0.92 / 0.86$ & $0.55 / 0.93 / 0.94$ & $0.30 / 0.96 / 0.90$ & $\underline{0.73} / 0.78 / \underline{0.95}$ & $0.52 / 0.94 / 0.89$ & $\underline{0.73} / 0.83 / \underline{0.72}$ & $0.59 / 0.83 / 0.78$ \\
Layer-wise $L_2$ & $0.65 / 0.88 / 0.92$ & $0.68 / 0.94 / \underline{0.97}$ & $0.66 / 0.91 / 0.97$ & $0.60 / 0.88 / 0.92$ & $\underline{0.65} / 0.94 / \underline{0.93}$ & $0.26 / 0.98 / 0.42$ & $0.38 / 1.00 / 0.69$ \\
Time-wise $L_2$ & $0.79 / 0.73 / \underline{0.96}$ & $0.82 / 0.68 / \mathbf{0.98}$ & $\underline{0.82} / 0.67 / \mathbf{0.99}$ & $0.65 / 0.88 / 0.93$ & $0.64 / 0.91 / \underline{0.93}$ & $0.33 / 0.94 / 0.44$ & $0.39 / 0.98 / 0.69$ \\
\textbf{StALT (Ours)} & $\mathbf{0.82} / \underline{0.61} / \mathbf{0.97}$ & $\mathbf{0.84} / \underline{0.57} / \mathbf{0.98}$ & $0.81 / \underline{0.62} / \underline{0.98}$ & $0.68 / 0.79 / 0.94$ & $0.64 / \underline{0.88} / \underline{0.93}$ & $0.59 / 0.88 / 0.63$ & $0.54 / 0.91 / 0.76$ \\
\midrule
\multicolumn{8}{c}{\textbf{GPQA-Diamond (AUROC $\uparrow$ / FPR95 $\downarrow$ / AUPR $\uparrow$)}} \\ \midrule
Gen. Tokens & $\underline{0.59} / \mathbf{0.91} / 0.42$ & $0.66 / 0.91 / 0.60$ & $0.68 / 0.92 / 0.67$ & $\mathbf{0.58} / \mathbf{0.90} / \mathbf{0.39}$ & $\mathbf{0.77} / \mathbf{0.72} / \mathbf{0.78}$ & $\mathbf{0.52} / 0.94 / \mathbf{0.29}$ & $\mathbf{0.54} / \mathbf{0.87} / \underline{0.30}$ \\
Max Prob. & $0.57 / 0.95 / 0.42$ & $0.62 / 0.96 / 0.61$ & $0.65 / 0.95 / 0.63$ & $0.54 / 0.94 / 0.37$ & $0.60 / 0.97 / 0.65$ & $0.49 / 0.94 / 0.27$ & $0.50 / 0.96 / 0.25$ \\
Perplexity & $0.43 / 0.98 / 0.32$ & $0.37 / 0.98 / 0.42$ & $0.35 / 0.98 / 0.43$ & $0.46 / 0.95 / 0.31$ & $\underline{0.65} / \underline{0.92} / \underline{0.70}$ & $\underline{0.51} / 0.94 / \underline{0.28}$ & $0.50 / \underline{0.89} / 0.27$ \\
Entropy & $0.57 / 0.95 / 0.42$ & $0.63 / 0.96 / 0.62$ & $0.65 / 0.95 / 0.64$ & $0.54 / 0.94 / 0.37$ & $0.57 / 0.97 / 0.63$ & $0.50 / 0.94 / 0.27$ & $0.49 / 0.96 / 0.25$ \\
CoE-R & $0.53 / 0.95 / 0.40$ & $0.50 / 0.98 / 0.50$ & $0.47 / 0.99 / 0.49$ & $0.48 / 0.95 / 0.33$ & $0.39 / 0.98 / 0.50$ & $0.50 / 0.97 / \mathbf{0.29}$ & $0.50 / 0.94 / 0.27$ \\
CoE-C & $0.53 / 0.95 / 0.41$ & $0.49 / 0.99 / 0.49$ & $0.47 / 0.99 / 0.49$ & $0.54 / 0.93 / 0.36$ & $0.39 / 0.98 / 0.50$ & $0.50 / 0.96 / \mathbf{0.29}$ & $0.52 / 0.96 / 0.28$ \\
Layer-wise $L_2$ & $0.54 / 0.94 / 0.38$ & $0.52 / 0.94 / 0.50$ & $0.57 / 0.92 / 0.56$ & $\underline{0.56} / 0.96 / \underline{0.38}$ & $0.62 / 0.95 / 0.67$ & $0.45 / 0.96 / 0.25$ & $\underline{0.53} / 0.94 / 0.29$ \\
Time-wise $L_2$ & $0.56 / 0.95 / 0.41$ & $0.64 / 0.93 / 0.60$ & $0.63 / 0.96 / 0.65$ & $0.50 / 0.95 / 0.33$ & $0.42 / 0.99 / 0.54$ & $0.45 / 0.94 / 0.25$ & $\mathbf{0.54} / 0.93 / 0.28$ \\
\textbf{StALT (Ours)} & $\mathbf{0.60} / \underline{0.92} / \mathbf{0.47}$ & $\mathbf{0.70} / \mathbf{0.86} / \mathbf{0.69}$ & $\mathbf{0.72} / \mathbf{0.87} / \underline{0.70}$ & $\underline{0.56} / 0.93 / 0.37$ & $0.60 / 0.93 / 0.65$ & $0.50 / 0.94 / 0.26$ & $\mathbf{0.54} / 0.95 / \underline{0.30}$ \\
\midrule
\multicolumn{8}{c}{\textbf{MMLU-Pro (STEM) (AUROC $\uparrow$ / FPR95 $\downarrow$ / AUPR $\uparrow$)}} \\ \midrule
Gen. Tokens & $0.78 / 0.71 / \underline{0.81}$ & $0.74 / 0.68 / 0.84$ & $0.73 / 0.70 / 0.82$ & $\mathbf{0.72} / \mathbf{0.75} / \mathbf{0.68}$ & $\mathbf{0.73} / \mathbf{0.71} / \mathbf{0.85}$ & $\mathbf{0.60} / \mathbf{0.92} / \underline{0.38}$ & $\mathbf{0.72} / \mathbf{0.83} / \mathbf{0.59}$ \\
Max Prob. & $0.72 / 0.90 / 0.78$ & $0.67 / 0.88 / 0.81$ & $0.65 / 0.89 / 0.78$ & $0.58 / 0.94 / 0.56$ & $0.54 / 0.97 / 0.73$ & $0.47 / 0.95 / 0.30$ & $0.52 / 0.94 / 0.33$ \\
Perplexity & $0.28 / 0.99 / 0.46$ & $0.32 / 0.98 / 0.59$ & $0.35 / 0.98 / 0.59$ & $0.41 / 0.97 / 0.44$ & $\underline{0.62} / \underline{0.91} / \underline{0.79}$ & $0.52 / \mathbf{0.92} / 0.33$ & $0.47 / 0.87 / 0.31$ \\
Entropy & $0.72 / 0.91 / 0.78$ & $0.68 / 0.88 / 0.82$ & $0.66 / 0.89 / 0.79$ & $0.60 / 0.93 / 0.57$ & $0.54 / 0.98 / 0.73$ & $0.47 / 0.94 / 0.30$ & $0.54 / 0.93 / 0.34$ \\
CoE-R & $0.39 / 0.97 / 0.50$ & $0.39 / 0.95 / 0.60$ & $0.41 / 0.95 / 0.61$ & $0.62 / 0.89 / 0.59$ & $0.39 / 0.95 / 0.63$ & $0.46 / 0.96 / 0.32$ & $\underline{0.57} / \underline{0.84} / \underline{0.36}$ \\
CoE-C & $0.42 / 0.97 / 0.53$ & $0.42 / 0.94 / 0.62$ & $0.42 / 0.95 / 0.62$ & $0.64 / 0.84 / 0.60$ & $0.39 / 0.95 / 0.63$ & $0.47 / 0.96 / 0.32$ & $0.52 / 0.92 / 0.33$ \\
Layer-wise $L_2$ & $0.57 / 0.94 / 0.65$ & $0.54 / 0.97 / 0.72$ & $0.50 / 0.97 / 0.70$ & $0.45 / 0.95 / 0.46$ & $0.58 / 0.97 / 0.75$ & $0.54 / 0.94 / 0.34$ & $0.43 / 0.98 / 0.29$ \\
Time-wise $L_2$ & $0.76 / 0.82 / \underline{0.81}$ & $0.74 / 0.83 / 0.84$ & $0.71 / 0.85 / 0.81$ & $0.56 / 0.95 / 0.54$ & $0.50 / 0.99 / 0.70$ & $\underline{0.59} / \underline{0.93} / \mathbf{0.39}$ & $0.46 / 0.95 / 0.32$ \\
\textbf{StALT (Ours)} & $\mathbf{0.80} / \mathbf{0.66} / \mathbf{0.82}$ & $\mathbf{0.79} / \mathbf{0.65} / \mathbf{0.86}$ & $\mathbf{0.77} / \mathbf{0.67} / \mathbf{0.84}$ & $\underline{0.65} / \underline{0.83} / \underline{0.61}$ & $0.53 / 0.93 / 0.72$ & $0.52 / 0.96 / 0.35$ & $0.56 / 0.90 / \underline{0.36}$ \\
\midrule
\multicolumn{8}{c}{\textbf{MMLU-Pro (Non-STEM) (AUROC $\uparrow$ / FPR95 $\downarrow$ / AUPR $\uparrow$)}} \\ \midrule
Gen. Tokens & $\mathbf{0.69} / 0.88 / \underline{0.61}$ & $\underline{0.68} / \underline{0.90} / 0.70$ & $0.68 / 0.87 / 0.71$ & $\mathbf{0.65} / \mathbf{0.88} / \mathbf{0.51}$ & $\mathbf{0.72} / \mathbf{0.85} / \mathbf{0.77}$ & $0.51 / \underline{0.93} / 0.29$ & $\mathbf{0.69} / \mathbf{0.89} / \mathbf{0.50}$ \\
Max Prob. & $\mathbf{0.69} / 0.93 / \mathbf{0.63}$ & $\mathbf{0.72} / 0.91 / \mathbf{0.74}$ & $\mathbf{0.71} / 0.88 / \mathbf{0.74}$ & $0.56 / 0.94 / 0.40$ & $0.52 / 0.98 / 0.61$ & $\underline{0.57} / \mathbf{0.92} / \underline{0.32}$ & $0.52 / \underline{0.94} / 0.35$ \\
Perplexity & $0.32 / 1.00 / 0.32$ & $0.28 / 1.00 / 0.40$ & $0.29 / 0.99 / 0.43$ & $0.44 / 0.95 / 0.32$ & $0.61 / 0.93 / \underline{0.69}$ & $0.42 / 0.95 / 0.24$ & $0.48 / 0.98 / 0.31$ \\
Entropy & $\mathbf{0.69} / 0.93 / \mathbf{0.63}$ & $\mathbf{0.72} / 0.91 / \mathbf{0.74}$ & $\mathbf{0.71} / 0.88 / \mathbf{0.74}$ & $0.57 / 0.93 / 0.40$ & $0.51 / 0.98 / 0.60$ & $\mathbf{0.58} / \mathbf{0.92} / \mathbf{0.33}$ & $\underline{0.53} / \underline{0.94} / \underline{0.36}$ \\
CoE-R & $0.50 / 0.96 / 0.41$ & $0.57 / 0.94 / 0.59$ & $0.55 / 0.94 / 0.60$ & $0.54 / \underline{0.92} / 0.38$ & $\underline{0.65} / \underline{0.87} / \underline{0.69}$ & $0.52 / 0.95 / \underline{0.32}$ & $0.46 / \underline{0.94} / 0.29$ \\
CoE-C & $0.52 / 0.95 / 0.42$ & $0.61 / 0.92 / 0.61$ & $0.56 / 0.93 / 0.61$ & $0.55 / 0.93 / 0.39$ & $\underline{0.65} / \underline{0.87} / \underline{0.69}$ & $0.52 / 0.95 / \underline{0.32}$ & $0.46 / \underline{0.94} / 0.29$ \\
Layer-wise $L_2$ & $0.60 / 0.95 / 0.49$ & $0.63 / 0.94 / 0.63$ & $0.52 / 0.94 / 0.58$ & $0.56 / 0.98 / 0.39$ & $0.62 / 0.96 / 0.68$ & $0.50 / 0.95 / 0.28$ & $0.52 / 0.99 / 0.33$ \\
Time-wise $L_2$ & $\mathbf{0.69} / 0.88 / 0.59$ & $0.66 / \underline{0.90} / 0.68$ & $0.66 / 0.89 / 0.69$ & $0.50 / 0.93 / 0.36$ & $0.50 / 0.98 / 0.57$ & $0.49 / 0.96 / 0.28$ & $0.45 / 0.98 / 0.30$ \\
\textbf{StALT (Ours)} & $\mathbf{0.69} / \underline{0.85} / 0.59$ & $\underline{0.68} / \mathbf{0.89} / 0.69$ & $\underline{0.70} / \mathbf{0.83} / 0.72$ & $\underline{0.60} / \underline{0.92} / \underline{0.45}$ & $0.56 / 0.92 / 0.61$ & $0.41 / 0.97 / 0.24$ & $0.48 / \underline{0.94} / 0.31$ \\
\bottomrule
\end{tabular}
}
\vspace{0.2em}
\caption{Correctness prediction across models and datasets. Bold denotes the best score and underline the second best score within each model column.}
\label{tab:comprehensive_results}
\end{table*}

\subsection{Interventions}
This appendix reports supplementary results for the intervention analyses in Section~\ref{sec:perturbation}.
Here we present additional results on GPQA-Diamond and Qwen3-8B to test the generality of the observed patterns.

\subsubsection{Reasoning-Amplifying Interventions on GPQA-Diamond}

\begin{figure}[t]
    \centering
    \includegraphics[width=0.6\textwidth]{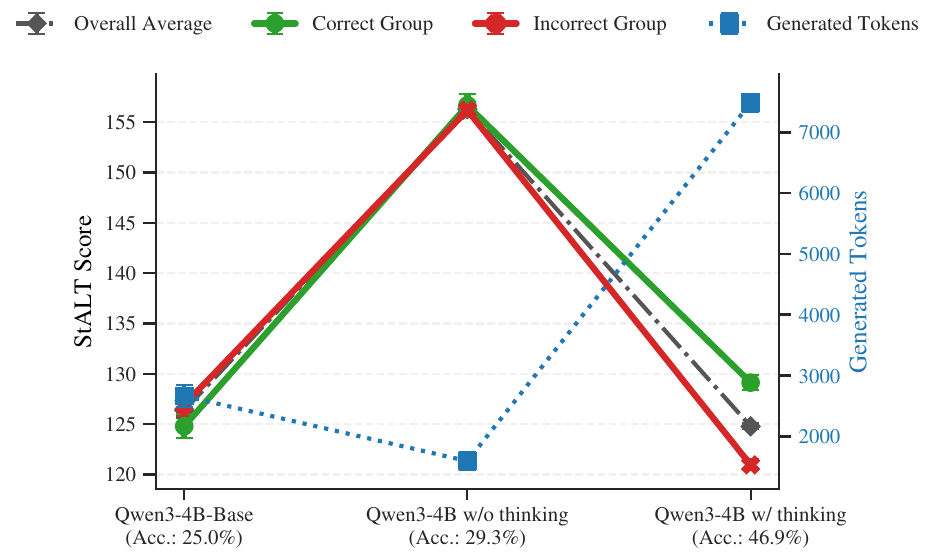}
    \caption{Within-family comparison on GPQA-Diamond for Qwen3-4B across base, no-thinking, and thinking-mode settings.}
    \label{fig:thinking_mode_gpqad_app}
\end{figure}

Figure~\ref{fig:thinking_mode_gpqad_app} shows the thinking-mode comparison on GPQA-Diamond, complementing the s1K-1.1 results in the main text (Figure~\ref{fig:thinking_mode}).
Unlike the monotonic increase observed on s1K-1.1, GPQA-Diamond reveals a non-monotonic pattern.
The no-thinking condition exhibits the largest \stalt{}, despite producing the shortest outputs of the three conditions.
Enabling thinking mode increases accuracy substantially and nearly triples the output length, yet \stalt{} drops back to the level of the base model.

One plausible explanation is that, unlike the purely algorithmic integer problems, GPQA-Diamond requires substantial domain knowledge in addition to reasoning.
With thinking mode enabled, the model produces a longer trace that mixes knowledge retrieval and option comparison with active reasoning.
Because \stalt{} is time-averaged, these knowledge-heavy tokens can dilute the per-step dynamical signature, consistent with the domain-dependent pattern in Section~\ref{sec:observations}, where knowledge-oriented tasks show weaker dynamics.
By contrast, in the no-thinking condition, reasoning-related computation appears compressed into fewer tokens, yielding larger per-step state updates and hence higher \stalt{}.
These results suggest that the effect of thinking mode on the hidden-state signature is task-dependent: it is strongest when the task is predominantly computational and can be diluted when the task also demands extensive domain-knowledge retrieval.

\subsubsection{In-Context CoT Scaffolding with Qwen3-8B}

\begin{figure}[t]
    \centering
    \includegraphics[width=0.6\textwidth]{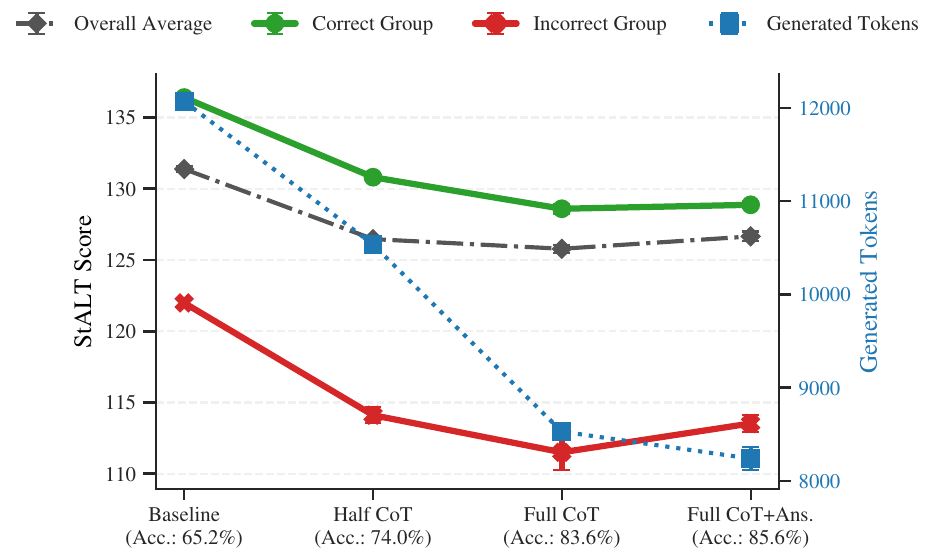}
    \caption{Effect of in-context CoT scaffolding on hidden-state dynamics for Qwen3-8B on integer problems from s1K-1.1.}
    \label{fig:cot_qwen38b_app}
\end{figure}

Figure~\ref{fig:cot_qwen38b_app} shows the CoT scaffolding experiment repeated with Qwen3-8B, complementing the Qwen3-4B results in Figure~\ref{fig:cot}.
The same qualitative pattern holds: as more of the solution is supplied externally, accuracy improves monotonically and generated length decreases.
Both the correct and incorrect groups individually show declining \stalt{} under increasing scaffolding, confirming that external reasoning substitutes for internal computation across model scales within the Qwen3 family.

\subsubsection{Supervised Fine-Tuning on GPQA-Diamond}

\begin{figure}[t]
    \centering
    \includegraphics[width=0.6\textwidth]{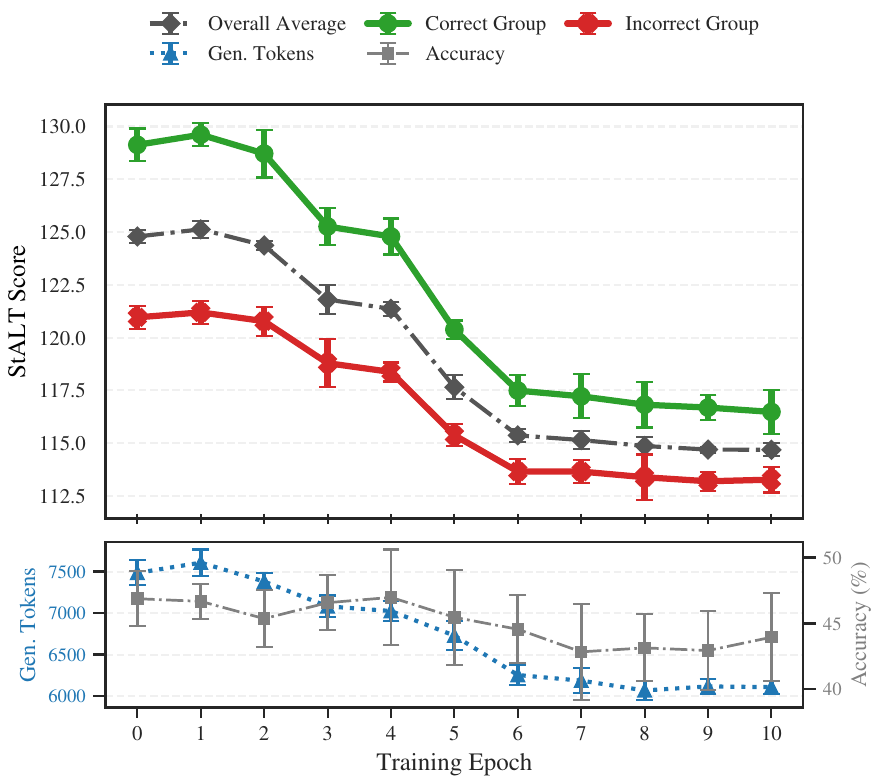}
    \caption{Effect of supervised fine-tuning (trained on s1K-1.1 integer problems) evaluated on GPQA-Diamond.}
    \label{fig:sft_gpqad_app}
\end{figure}

Figure~\ref{fig:sft_gpqad_app} shows the out-of-domain effect of the same fine-tuning procedure (trained on s1K-1.1 integer problems) when evaluated on GPQA-Diamond, complementing the in-domain results in Figure~\ref{fig:sft}.
Here, \stalt{} declines over training checkpoints, but unlike the in-domain setting, accuracy also declines rather than improving.
Mean generated length decreases in parallel.
This combination suggests that the fine-tuning procedure that helps on training-aligned integer problems simultaneously weakens the broader reasoning behavior needed for GPQA-Diamond.
Rather than indicating a universal gain in reasoning efficiency, the fine-tuning intervention appears to trade transferable internal computation for more specialized behavior on the fine-tuning domain.

\section{Ablations}
\label{app:metric_ablations}
This section provides additional ablations of \stalt{} to isolate which parts of the metric are responsible for its signal.
We organize the analysis into three groups.
First, component and aggregation ablations examine the role of the aggregation order and the choice of similarity function.
Second, segment and truncation ablations test which portion of the trajectory carries the signal and how much of it is needed for the signal to emerge.
Third, a softmax temperature sweep examines how sensitive \stalt{} is to the selectivity of the layer-local weighting.

\subsection{Component and aggregation ablations}
\label{app:component_aggregation_ablations}

\begin{table*}[t]
\centering
\resizebox{\textwidth}{!}{
\begin{tabular}{l|ccccc|cc}
\toprule
\textbf{Variant / Model} & \textbf{Qwen3-1.7B} & \textbf{Qwen3-4B} & \textbf{Qwen3-8B} & \textbf{SmolLM3-3B} & \textbf{gpt-oss-20b} & \textbf{Qwen2.5-Math-7B} & \textbf{Llama-3.1-8B-Instruct} \\ \midrule
\midrule
\multicolumn{8}{c}{\textbf{MATH-500 (AUROC / FPR95 / AUPR)}} \\
\midrule
StALT & $0.78 / 0.58 / 0.92$ & $0.69 / 0.66 / 0.92$ & $0.74 / 0.69 / 0.94$ & $0.72 / 0.67 / 0.91$ & $0.60 / 0.86 / 0.85$ & $0.54 / 0.84 / 0.42$ & $0.56 / 0.87 / 0.45$ \\
Time-wise $L_2$ & $0.72 / 0.74 / 0.88$ & $0.69 / 0.80 / 0.92$ & $0.67 / 0.83 / 0.91$ & $0.63 / 0.92 / 0.87$ & $0.55 / 0.95 / 0.84$ & $0.37 / 0.93 / 0.34$ & $0.40 / 0.96 / 0.36$ \\
Layer-wise $L_2$ & $0.59 / 0.89 / 0.84$ & $0.56 / 0.94 / 0.87$ & $0.51 / 0.97 / 0.86$ & $0.41 / 0.95 / 0.79$ & $0.54 / 0.97 / 0.83$ & $0.34 / 0.94 / 0.33$ & $0.35 / 0.99 / 0.34$ \\
\addlinespace[0.35em]
\multicolumn{8}{l}{\textit{Aggregation order}} \\
Reversed StALT & $0.35 / 1.00 / 0.73$ & $0.39 / 1.00 / 0.82$ & $0.40 / 1.00 / 0.83$ & $0.39 / 0.98 / 0.77$ & $0.52 / 0.96 / 0.83$ & $0.35 / 0.95 / 0.32$ & $0.66 / 0.85 / 0.57$ \\
\addlinespace[0.35em]
\multicolumn{8}{l}{\textit{Cosine similarity}} \\
Time-wise Cosine & $0.33 / 0.99 / 0.72$ & $0.36 / 0.97 / 0.80$ & $0.36 / 0.98 / 0.80$ & $0.44 / 0.94 / 0.80$ & $0.45 / 0.98 / 0.80$ & $0.69 / 0.65 / 0.54$ & $0.68 / 0.84 / 0.56$ \\
Layer-wise Cosine & $0.50 / 1.00 / 0.80$ & $0.49 / 1.00 / 0.86$ & $0.49 / 1.00 / 0.85$ & $0.59 / 0.92 / 0.85$ & $0.41 / 0.99 / 0.77$ & $0.77 / 0.59 / 0.62$ & $0.71 / 0.83 / 0.58$ \\
\midrule
\multicolumn{8}{c}{\textbf{GSM8K (AUROC / FPR95 / AUPR)}} \\
\midrule
StALT & $0.82 / 0.61 / 0.97$ & $0.84 / 0.57 / 0.98$ & $0.81 / 0.62 / 0.98$ & $0.68 / 0.79 / 0.94$ & $0.64 / 0.88 / 0.93$ & $0.59 / 0.88 / 0.63$ & $0.54 / 0.91 / 0.76$ \\
Time-wise $L_2$ & $0.79 / 0.73 / 0.96$ & $0.82 / 0.68 / 0.98$ & $0.82 / 0.67 / 0.99$ & $0.65 / 0.88 / 0.93$ & $0.64 / 0.91 / 0.93$ & $0.33 / 0.94 / 0.44$ & $0.39 / 0.98 / 0.69$ \\
Layer-wise $L_2$ & $0.65 / 0.88 / 0.92$ & $0.68 / 0.94 / 0.97$ & $0.66 / 0.91 / 0.97$ & $0.60 / 0.88 / 0.92$ & $0.65 / 0.94 / 0.93$ & $0.26 / 0.98 / 0.42$ & $0.38 / 1.00 / 0.69$ \\
\addlinespace[0.35em]
\multicolumn{8}{l}{\textit{Aggregation order}} \\
Reversed StALT & $0.35 / 0.99 / 0.85$ & $0.21 / 1.00 / 0.87$ & $0.23 / 1.00 / 0.89$ & $0.45 / 0.97 / 0.88$ & $0.62 / 0.86 / 0.92$ & $0.31 / 0.93 / 0.42$ & $0.62 / 0.83 / 0.79$ \\
\addlinespace[0.35em]
\multicolumn{8}{l}{\textit{Cosine similarity}} \\
Time-wise Cosine & $0.19 / 1.00 / 0.80$ & $0.17 / 1.00 / 0.87$ & $0.16 / 1.00 / 0.89$ & $0.34 / 0.97 / 0.84$ & $0.37 / 0.99 / 0.86$ & $0.67 / 0.67 / 0.63$ & $0.65 / 0.89 / 0.82$ \\
Layer-wise Cosine & $0.34 / 1.00 / 0.86$ & $0.30 / 0.99 / 0.91$ & $0.33 / 1.00 / 0.93$ & $0.52 / 0.95 / 0.90$ & $0.36 / 0.98 / 0.86$ & $0.80 / 0.48 / 0.77$ & $0.57 / 0.90 / 0.77$ \\
\midrule
\multicolumn{8}{c}{\textbf{GPQA-Diamond (AUROC / FPR95 / AUPR)}} \\
\midrule
StALT & $0.60 / 0.92 / 0.47$ & $0.70 / 0.86 / 0.69$ & $0.72 / 0.87 / 0.70$ & $0.56 / 0.93 / 0.37$ & $0.60 / 0.93 / 0.65$ & $0.50 / 0.94 / 0.26$ & $0.54 / 0.95 / 0.30$ \\
Time-wise $L_2$ & $0.56 / 0.95 / 0.41$ & $0.64 / 0.93 / 0.60$ & $0.63 / 0.96 / 0.65$ & $0.50 / 0.95 / 0.33$ & $0.42 / 0.99 / 0.54$ & $0.45 / 0.94 / 0.25$ & $0.54 / 0.93 / 0.28$ \\
Layer-wise $L_2$ & $0.54 / 0.94 / 0.38$ & $0.52 / 0.94 / 0.50$ & $0.57 / 0.92 / 0.56$ & $0.56 / 0.96 / 0.38$ & $0.62 / 0.95 / 0.67$ & $0.45 / 0.96 / 0.25$ & $0.53 / 0.94 / 0.29$ \\
\addlinespace[0.35em]
\multicolumn{8}{l}{\textit{Aggregation order}} \\
Reversed StALT & $0.49 / 0.96 / 0.34$ & $0.54 / 0.96 / 0.53$ & $0.62 / 0.98 / 0.61$ & $0.50 / 0.96 / 0.33$ & $0.74 / 0.75 / 0.75$ & $0.46 / 0.98 / 0.27$ & $0.49 / 0.96 / 0.28$ \\
\addlinespace[0.35em]
\multicolumn{8}{l}{\textit{Cosine similarity}} \\
Time-wise Cosine & $0.44 / 0.97 / 0.32$ & $0.43 / 0.96 / 0.43$ & $0.42 / 0.99 / 0.46$ & $0.42 / 0.95 / 0.30$ & $0.64 / 0.97 / 0.70$ & $0.56 / 0.89 / 0.32$ & $0.47 / 0.98 / 0.27$ \\
Layer-wise Cosine & $0.44 / 0.99 / 0.32$ & $0.47 / 0.99 / 0.46$ & $0.46 / 0.99 / 0.48$ & $0.45 / 0.99 / 0.31$ & $0.24 / 1.00 / 0.46$ & $0.56 / 0.95 / 0.31$ & $0.51 / 0.97 / 0.27$ \\
\midrule
\multicolumn{8}{c}{\textbf{MMLU-Pro (STEM) (AUROC / FPR95 / AUPR)}} \\
\midrule
StALT & $0.80 / 0.66 / 0.82$ & $0.79 / 0.65 / 0.86$ & $0.77 / 0.67 / 0.84$ & $0.65 / 0.83 / 0.61$ & $0.53 / 0.93 / 0.72$ & $0.52 / 0.96 / 0.35$ & $0.56 / 0.90 / 0.36$ \\
Time-wise $L_2$ & $0.76 / 0.82 / 0.81$ & $0.74 / 0.83 / 0.84$ & $0.71 / 0.85 / 0.81$ & $0.56 / 0.95 / 0.54$ & $0.50 / 0.99 / 0.70$ & $0.59 / 0.93 / 0.39$ & $0.46 / 0.95 / 0.32$ \\
Layer-wise $L_2$ & $0.57 / 0.94 / 0.65$ & $0.54 / 0.97 / 0.72$ & $0.50 / 0.97 / 0.70$ & $0.45 / 0.95 / 0.46$ & $0.58 / 0.97 / 0.75$ & $0.54 / 0.94 / 0.34$ & $0.43 / 0.98 / 0.29$ \\
\addlinespace[0.35em]
\multicolumn{8}{l}{\textit{Aggregation order}} \\
Reversed StALT & $0.35 / 0.99 / 0.48$ & $0.38 / 0.98 / 0.61$ & $0.38 / 0.97 / 0.61$ & $0.40 / 0.97 / 0.41$ & $0.60 / 0.91 / 0.77$ & $0.42 / 0.96 / 0.28$ & $0.59 / 0.90 / 0.39$ \\
\addlinespace[0.35em]
\multicolumn{8}{l}{\textit{Cosine similarity}} \\
Time-wise Cosine & $0.27 / 0.99 / 0.46$ & $0.32 / 0.98 / 0.59$ & $0.32 / 0.98 / 0.58$ & $0.36 / 0.98 / 0.41$ & $0.56 / 0.96 / 0.75$ & $0.47 / 0.98 / 0.31$ & $0.57 / 0.93 / 0.38$ \\
Layer-wise Cosine & $0.41 / 0.99 / 0.54$ & $0.43 / 1.00 / 0.66$ & $0.45 / 0.98 / 0.66$ & $0.54 / 0.99 / 0.52$ & $0.31 / 0.99 / 0.63$ & $0.54 / 0.95 / 0.36$ & $0.59 / 0.93 / 0.38$ \\
\midrule
\multicolumn{8}{c}{\textbf{MMLU-Pro (Non-STEM) (AUROC / FPR95 / AUPR)}} \\
\midrule
StALT & $0.69 / 0.85 / 0.59$ & $0.68 / 0.89 / 0.69$ & $0.70 / 0.83 / 0.72$ & $0.60 / 0.92 / 0.45$ & $0.56 / 0.92 / 0.61$ & $0.41 / 0.97 / 0.24$ & $0.48 / 0.94 / 0.31$ \\
Time-wise $L_2$ & $0.69 / 0.88 / 0.59$ & $0.66 / 0.90 / 0.68$ & $0.66 / 0.89 / 0.69$ & $0.50 / 0.93 / 0.36$ & $0.50 / 0.98 / 0.57$ & $0.49 / 0.96 / 0.28$ & $0.45 / 0.98 / 0.30$ \\
Layer-wise $L_2$ & $0.60 / 0.95 / 0.49$ & $0.63 / 0.94 / 0.63$ & $0.52 / 0.94 / 0.58$ & $0.56 / 0.98 / 0.39$ & $0.62 / 0.96 / 0.68$ & $0.50 / 0.95 / 0.28$ & $0.52 / 0.99 / 0.33$ \\
\addlinespace[0.35em]
\multicolumn{8}{l}{\textit{Aggregation order}} \\
Reversed StALT & $0.44 / 0.98 / 0.37$ & $0.46 / 0.98 / 0.53$ & $0.41 / 0.98 / 0.52$ & $0.46 / 0.96 / 0.32$ & $0.64 / 0.90 / 0.70$ & $0.53 / 0.95 / 0.29$ & $0.47 / 0.94 / 0.30$ \\
\addlinespace[0.35em]
\multicolumn{8}{l}{\textit{Cosine similarity}} \\
Time-wise Cosine & $0.36 / 0.99 / 0.34$ & $0.38 / 0.99 / 0.47$ & $0.37 / 0.98 / 0.48$ & $0.39 / 0.98 / 0.30$ & $0.54 / 0.95 / 0.61$ & $0.55 / 0.95 / 0.33$ & $0.49 / 0.97 / 0.31$ \\
Layer-wise Cosine & $0.33 / 1.00 / 0.33$ & $0.32 / 1.00 / 0.45$ & $0.37 / 1.00 / 0.50$ & $0.46 / 0.97 / 0.34$ & $0.32 / 0.99 / 0.49$ & $0.57 / 0.95 / 0.34$ & $0.48 / 0.96 / 0.31$ \\
\bottomrule
\end{tabular}
}
\vspace{0.2em}
\caption{Component and aggregation ablations for StALT across datasets and model families. Rows compare the full StALT and $L_2$ component scores against reversed aggregation order and cosine-similarity variants.}
\label{tab:component_aggregation_ablations}
\end{table*}

\subsubsection{Aggregation order.}
\label{app:aggregation_order}

The standard \stalt{} metric treats temporal change as the primary signal, utilizing layer-local variation to weight the contribution of each layer. 
To evaluate the necessity of this design, we introduce a \textit{reversed \stalt{}} variant (Table~\ref{tab:component_aggregation_ablations}), which swaps these roles by using time-wise saliency to weight layer-local changes. 

Across all five datasets, \stalt{} consistently outperforms its reversed counterpart on reasoning models. 
This performance gap is particularly pronounced for the Qwen3 family on MATH-500 and GSM8K, where reversed \stalt{} frequently degrades to near-random performance, whereas \stalt{} maintains robust discriminative power. 

Conversely, on non-reasoning baselines, reversed \stalt{} can match or even surpass \stalt{}. 
This suggests that in the absence of a reasoning-specific temporal signature, layer-local variations may primarily capture residual structural patterns. 

Overall, the comparison confirms the asymmetry suggested by the main results: the hidden-state pattern is better characterized as temporally driven computation with non-uniform layer concentration than as layer-local change selected over time.

\subsubsection{Cosine Similarity}

We also test cosine-similarity variants of the two primitive quantities, replacing the $L_2$ distance with cosine similarity to measure directional change across time or layers.
As shown in Table~\ref{tab:component_aggregation_ablations}, both cosine variants are generally much weaker than the $L_2$-based formulation on reasoning models.
On the Qwen3 family, cosine-based scores fall substantially below \stalt{} across all reasoning-intensive datasets, with the time-wise cosine variant often dropping below chance level.
This gap suggests that the correctness-associated signal lies primarily in the magnitude of hidden-state movement rather than in its direction alone.

The cosine variants are not uniformly uninformative, however.
On non-reasoning or task-specialized baselines such as Qwen2.5-Math-7B and Llama-3.1-8B-Instruct, cosine-based scores can become competitive with or exceed \stalt{}.
This boundary case suggests that when the reasoning-specific dynamical signature is absent, coarse directional regularities may still correlate with correctness, but they do not capture the magnitude-driven, layer-localized variation that characterizes reasoning models.
Overall, these results justify the use of $L_2$ distance as the building block for \stalt{}.

\subsection{Segment and truncation ablations}
\label{app:segment_truncation_ablations}

\begin{table*}[t]
\centering
\resizebox{\textwidth}{!}{
\begin{tabular}{l|ccccc|cc}
\toprule
\textbf{Variant / Model} & \textbf{Qwen3-1.7B} & \textbf{Qwen3-4B} & \textbf{Qwen3-8B} & \textbf{SmolLM3-3B} & \textbf{gpt-oss-20b} & \textbf{Qwen2.5-Math-7B} & \textbf{Llama-3.1-8B-Instruct} \\ \midrule
\midrule
\multicolumn{8}{c}{\textbf{MATH-500 (AUROC / FPR95 / AUPR)}} \\
\midrule
StALT full trace & $0.78 / 0.58 / 0.92$ & $0.69 / 0.66 / 0.92$ & $0.74 / 0.69 / 0.94$ & $0.72 / 0.67 / 0.91$ & $0.60 / 0.86 / 0.85$ & $0.54 / 0.84 / 0.42$ & $0.56 / 0.87 / 0.45$ \\
\addlinespace[0.35em]
\multicolumn{8}{l}{\textit{Reasoning / answer tokens}} \\
StALT thinking only & $0.78 / 0.56 / 0.92$ & $0.70 / 0.66 / 0.92$ & $0.75 / 0.68 / 0.94$ & $0.72 / 0.69 / 0.91$ & $0.58 / 0.90 / 0.84$ & - / - / - & - / - / - \\
StALT answer only & $0.63 / 0.82 / 0.89$ & $0.55 / 0.85 / 0.90$ & $0.57 / 0.88 / 0.90$ & $0.57 / 0.86 / 0.87$ & $0.52 / 0.94 / 0.83$ & - / - / - & - / - / - \\
\addlinespace[0.35em]
\multicolumn{8}{l}{\textit{Early truncation}} \\
StALT first 25\% tokens & $0.69 / 0.82 / 0.88$ & $0.63 / 0.84 / 0.89$ & $0.69 / 0.82 / 0.92$ & $0.70 / 0.77 / 0.90$ & $0.57 / 0.90 / 0.85$ & $0.43 / 0.93 / 0.36$ & $0.54 / 0.91 / 0.44$ \\
StALT first 50\% tokens & $0.75 / 0.73 / 0.91$ & $0.67 / 0.78 / 0.91$ & $0.73 / 0.79 / 0.94$ & $0.71 / 0.78 / 0.91$ & $0.59 / 0.88 / 0.85$ & $0.42 / 0.91 / 0.35$ & $0.56 / 0.88 / 0.45$ \\
StALT first 75\% tokens & $0.77 / 0.64 / 0.92$ & $0.69 / 0.71 / 0.92$ & $0.75 / 0.72 / 0.94$ & $0.71 / 0.73 / 0.91$ & $0.60 / 0.86 / 0.86$ & $0.50 / 0.83 / 0.40$ & $0.56 / 0.89 / 0.45$ \\
\midrule
\multicolumn{8}{c}{\textbf{GSM8K (AUROC / FPR95 / AUPR)}} \\
\midrule
StALT full trace & $0.82 / 0.61 / 0.97$ & $0.84 / 0.57 / 0.98$ & $0.81 / 0.62 / 0.98$ & $0.68 / 0.79 / 0.94$ & $0.64 / 0.88 / 0.93$ & $0.59 / 0.88 / 0.63$ & $0.54 / 0.91 / 0.76$ \\
\addlinespace[0.35em]
\multicolumn{8}{l}{\textit{Reasoning / answer tokens}} \\
StALT thinking only & $0.82 / 0.62 / 0.97$ & $0.83 / 0.57 / 0.98$ & $0.81 / 0.62 / 0.98$ & $0.68 / 0.83 / 0.94$ & $0.63 / 0.91 / 0.93$ & - / - / - & - / - / - \\
StALT answer only & $0.81 / 0.67 / 0.97$ & $0.83 / 0.60 / 0.98$ & $0.82 / 0.66 / 0.99$ & $0.62 / 0.90 / 0.93$ & $0.63 / 0.88 / 0.93$ & - / - / - & - / - / - \\
\addlinespace[0.35em]
\multicolumn{8}{l}{\textit{Early truncation}} \\
StALT first 25\% tokens & $0.77 / 0.68 / 0.96$ & $0.77 / 0.64 / 0.97$ & $0.72 / 0.76 / 0.98$ & $0.64 / 0.82 / 0.93$ & $0.63 / 0.94 / 0.93$ & $0.56 / 0.90 / 0.58$ & $0.51 / 0.96 / 0.76$ \\
StALT first 50\% tokens & $0.81 / 0.67 / 0.97$ & $0.82 / 0.62 / 0.98$ & $0.80 / 0.66 / 0.98$ & $0.67 / 0.81 / 0.93$ & $0.65 / 0.90 / 0.94$ & $0.52 / 0.91 / 0.55$ & $0.53 / 0.95 / 0.76$ \\
StALT first 75\% tokens & $0.82 / 0.64 / 0.97$ & $0.83 / 0.60 / 0.98$ & $0.81 / 0.65 / 0.98$ & $0.68 / 0.83 / 0.94$ & $0.64 / 0.90 / 0.93$ & $0.59 / 0.85 / 0.60$ & $0.53 / 0.93 / 0.75$ \\
\midrule
\multicolumn{8}{c}{\textbf{GPQA-Diamond (AUROC / FPR95 / AUPR)}} \\
\midrule
StALT full trace & $0.60 / 0.92 / 0.47$ & $0.70 / 0.86 / 0.69$ & $0.72 / 0.87 / 0.70$ & $0.56 / 0.93 / 0.37$ & $0.60 / 0.93 / 0.65$ & $0.50 / 0.94 / 0.26$ & $0.54 / 0.95 / 0.30$ \\
\addlinespace[0.35em]
\multicolumn{8}{l}{\textit{Reasoning / answer tokens}} \\
StALT thinking only & $0.60 / 0.92 / 0.47$ & $0.70 / 0.88 / 0.70$ & $0.72 / 0.87 / 0.71$ & $0.55 / 0.93 / 0.37$ & $0.59 / 0.92 / 0.64$ & - / - / - & - / - / - \\
StALT answer only & $0.58 / 0.94 / 0.45$ & $0.62 / 0.87 / 0.57$ & $0.60 / 0.92 / 0.58$ & $0.55 / 0.94 / 0.37$ & $0.54 / 0.89 / 0.60$ & - / - / - & - / - / - \\
\addlinespace[0.35em]
\multicolumn{8}{l}{\textit{Early truncation}} \\
StALT first 25\% tokens & $0.58 / 0.93 / 0.42$ & $0.67 / 0.91 / 0.64$ & $0.69 / 0.88 / 0.69$ & $0.55 / 0.92 / 0.37$ & $0.55 / 0.98 / 0.62$ & $0.51 / 0.93 / 0.28$ & $0.52 / 0.92 / 0.28$ \\
StALT first 50\% tokens & $0.59 / 0.91 / 0.45$ & $0.68 / 0.89 / 0.67$ & $0.71 / 0.88 / 0.70$ & $0.56 / 0.93 / 0.37$ & $0.56 / 0.97 / 0.62$ & $0.51 / 0.91 / 0.27$ & $0.54 / 0.92 / 0.31$ \\
StALT first 75\% tokens & $0.60 / 0.92 / 0.47$ & $0.70 / 0.88 / 0.69$ & $0.72 / 0.88 / 0.71$ & $0.55 / 0.93 / 0.37$ & $0.59 / 0.95 / 0.64$ & $0.50 / 0.92 / 0.27$ & $0.54 / 0.94 / 0.31$ \\
\midrule
\multicolumn{8}{c}{\textbf{MMLU-Pro (STEM) (AUROC / FPR95 / AUPR)}} \\
\midrule
StALT full trace & $0.80 / 0.66 / 0.82$ & $0.79 / 0.65 / 0.86$ & $0.77 / 0.67 / 0.84$ & $0.65 / 0.83 / 0.61$ & $0.53 / 0.93 / 0.72$ & $0.52 / 0.96 / 0.35$ & $0.56 / 0.90 / 0.36$ \\
\addlinespace[0.35em]
\multicolumn{8}{l}{\textit{Reasoning / answer tokens}} \\
StALT thinking only & $0.80 / 0.67 / 0.82$ & $0.79 / 0.66 / 0.87$ & $0.77 / 0.67 / 0.85$ & $0.64 / 0.84 / 0.60$ & $0.52 / 0.95 / 0.72$ & - / - / - & - / - / - \\
StALT answer only & $0.73 / 0.74 / 0.75$ & $0.70 / 0.75 / 0.83$ & $0.69 / 0.76 / 0.80$ & $0.62 / 0.85 / 0.60$ & $0.56 / 0.90 / 0.74$ & - / - / - & - / - / - \\
\addlinespace[0.35em]
\multicolumn{8}{l}{\textit{Early truncation}} \\
StALT first 25\% tokens & $0.75 / 0.72 / 0.77$ & $0.76 / 0.74 / 0.85$ & $0.73 / 0.78 / 0.83$ & $0.63 / 0.87 / 0.59$ & $0.47 / 0.97 / 0.69$ & $0.50 / 0.98 / 0.35$ & $0.51 / 0.95 / 0.34$ \\
StALT first 50\% tokens & $0.79 / 0.69 / 0.81$ & $0.78 / 0.70 / 0.86$ & $0.75 / 0.73 / 0.84$ & $0.64 / 0.86 / 0.60$ & $0.49 / 0.96 / 0.70$ & $0.50 / 0.98 / 0.35$ & $0.53 / 0.93 / 0.35$ \\
StALT first 75\% tokens & $0.80 / 0.67 / 0.82$ & $0.79 / 0.67 / 0.86$ & $0.77 / 0.69 / 0.84$ & $0.64 / 0.84 / 0.60$ & $0.51 / 0.95 / 0.71$ & $0.51 / 0.97 / 0.35$ & $0.53 / 0.91 / 0.35$ \\
\midrule
\multicolumn{8}{c}{\textbf{MMLU-Pro (Non-STEM) (AUROC / FPR95 / AUPR)}} \\
\midrule
StALT full trace & $0.69 / 0.85 / 0.59$ & $0.68 / 0.89 / 0.69$ & $0.70 / 0.83 / 0.72$ & $0.60 / 0.92 / 0.45$ & $0.56 / 0.92 / 0.61$ & $0.41 / 0.97 / 0.24$ & $0.48 / 0.94 / 0.31$ \\
\addlinespace[0.35em]
\multicolumn{8}{l}{\textit{Reasoning / answer tokens}} \\
StALT thinking only & $0.69 / 0.85 / 0.60$ & $0.67 / 0.89 / 0.69$ & $0.70 / 0.83 / 0.73$ & $0.59 / 0.92 / 0.45$ & $0.57 / 0.92 / 0.63$ & - / - / - & - / - / - \\
StALT answer only & $0.67 / 0.86 / 0.57$ & $0.67 / 0.87 / 0.67$ & $0.67 / 0.85 / 0.69$ & $0.58 / 0.91 / 0.42$ & $0.59 / 0.91 / 0.63$ & - / - / - & - / - / - \\
\addlinespace[0.35em]
\multicolumn{8}{l}{\textit{Early truncation}} \\
StALT first 25\% tokens & $0.63 / 0.85 / 0.49$ & $0.65 / 0.89 / 0.65$ & $0.68 / 0.84 / 0.71$ & $0.58 / 0.92 / 0.42$ & $0.53 / 0.95 / 0.59$ & $0.43 / 0.97 / 0.26$ & $0.46 / 0.98 / 0.31$ \\
StALT first 50\% tokens & $0.68 / 0.84 / 0.56$ & $0.67 / 0.89 / 0.68$ & $0.70 / 0.83 / 0.72$ & $0.59 / 0.92 / 0.45$ & $0.54 / 0.93 / 0.59$ & $0.41 / 0.98 / 0.24$ & $0.47 / 0.97 / 0.31$ \\
StALT first 75\% tokens & $0.69 / 0.85 / 0.59$ & $0.67 / 0.89 / 0.69$ & $0.70 / 0.83 / 0.73$ & $0.59 / 0.92 / 0.45$ & $0.55 / 0.93 / 0.60$ & $0.41 / 0.98 / 0.24$ & $0.47 / 0.96 / 0.31$ \\
\bottomrule
\end{tabular}
}
\vspace{0.2em}
\caption{Segment and truncation ablations for StALT across datasets and model families. Rows compare the full-trace StALT score against thinking-only, answer-only and early-truncation variants.}
\label{tab:segment_truncation_ablations}
\end{table*}

\subsubsection{Reasoning vs.\ Answer Tokens}

For reasoning models that produce an explicit thinking trace followed by a final answer, we can isolate the contribution of each segment by computing \stalt{} over the tokens from the thinking section or the answer section only (Table~\ref{tab:segment_truncation_ablations}).
For Qwen3 and SmolLM3, we define the thinking section as the tokens enclosed by the \texttt{<think>} and \texttt{</think>} tags, and the answer section as the tokens generated after the closing \texttt{</think>} tag.
For gpt-oss-20b, we instead use the model's channel annotations, treating tokens in the \texttt{analysis} channel as the thinking section and tokens in the \texttt{final} channel as the answer section.
This segment-wise decomposition is available only for LRMs, and thus non-reasoning models are omitted from this comparison.

Across datasets, the thinking-only variant closely matches the full-trace \stalt{}, indicating that the predictive signal resides predominantly in the reasoning phase rather than the final-answer generation.
The answer-only variant consistently yields lower discrimination, with the gap being especially pronounced on datasets that demand sustained multi-step reasoning such as MATH-500 and GPQA-Diamond.
On MMLU-Pro Non-STEM, where the reasoning trace is shorter and the task relies more on knowledge retrieval, the difference between thinking-only and answer-only narrows.
These results confirm that the spatiotemporal signature captured by \stalt{} is primarily driven by the extended deliberation during the thinking phase, consistent with the interpretation that \stalt{} tracks internal computation rather than surface-level answer formatting.

\subsubsection{Early Truncation}

To test whether the signature requires the full reasoning trace or is already visible earlier in generation, we evaluate \stalt{} on the first 25\%, 50\%, or 75\% of its generated tokens.
Table~\ref{tab:segment_truncation_ablations} shows that performance consistently improves across datasets as more of the trajectory is observed, and that using the first 75\% of tokens is usually close to the full-trace result.
This pattern indicates that the signal is not confined to the very end of generation, but also accumulates over the later stages of the trajectory, which remain important for the strongest discrimination.

The early truncation results therefore suggest that \stalt{} already becomes informative before generation is complete, which is favorable for potential online monitoring applications.
At the same time, the best performance still tends to require most of the trace, consistent with our interpretation that successful reasoning is reflected in extended spatiotemporal dynamics rather than in a single localized burst.
On baselines where the full metric is already weak, truncation changes little in absolute terms, further suggesting that the main effect is specific to reasoning-oriented trajectories.

\subsection{Softmax temperature sweep}
\label{app:temperature_sweep}
\stalt{} computes a temporally averaged convex combination of temporal changes,
where the layer weights are determined by layer-local saliency.
The temperature parameter $\tau$ controls the sharpness of this layer selection: smaller values concentrate the weight on a small number of highly active layers, whereas larger values spread the weight more uniformly across layers.
As $\tau \to 0$, \stalt{} approaches the temporal amplitude of the most layer-active depth at each decoding step.
As $\tau \to \infty$, the layer weights approach a uniform distribution, and \stalt{} reduces to the unweighted average of temporal amplitude across layers.

We sweep the temperature parameter $\tau$, which controls how selectively layer-local variation is converted into weights.
In addition to finite temperature values, we include two limiting endpoints: $\tau\to 0$, corresponding to hard layer selection, and $\tau\to\infty$, corresponding to uniform layer weighting.
Table~\ref{tab:tau_sweep} reports AUROC for these settings.

Across most model-dataset combinations, performance is highly stable over the finite-temperature region.
The default setting $\tau=1$ remains very close to the best value in all reported settings.
In several Qwen3 settings, $\tau=100$ yields the best value, but the improvement over $\tau=1$ is small, indicating that the exact temperature value is not critical once the weighting remains layer-selective.

The uniform endpoint provides a useful contrast.
When $\tau\to\infty$, the layer weights become uniform and the layer-local selection mechanism is removed.
This suggests that StALT is not merely an unweighted temporal-amplitude statistic, and thus retaining layer-local selectivity contributes useful information.

Overall, the results indicate that StALT is robust to the precise choice of $\tau$ within the selective regime, while completely uniform layer weighting can discard useful layer-local structure.

\begin{table*}[t]
\centering
\resizebox{\textwidth}{!}{
\begin{tabular}{lccccccccccc}
\toprule
\textbf{Setting} & \textbf{$\tau\to 0$} & \textbf{$\tau=0.01$} & \textbf{$\tau=0.1$} & \textbf{$\tau=0.5$} & \textbf{$\tau=1$} & \textbf{$\tau=2$} & \textbf{$\tau=5$} & \textbf{$\tau=10$} & \textbf{$\tau=100$} & \textbf{$\tau\to\infty$} & \textbf{$|\Delta(\tau=1)|$ to best} \\
\midrule
Qwen3-1.7B / MATH-500 & $0.7771$ & $0.7771$ & $0.7771$ & $0.7771$ & $0.7771$ & $0.7771$ & $0.7771$ & $0.7771$ & $0.7772$ & $0.7248$ & $0.0001$ \\
Qwen3-4B / MATH-500 & $0.6901$ & $0.6901$ & $0.6901$ & $0.6901$ & $0.6901$ & $0.6901$ & $0.6901$ & $0.6902$ & $0.6930$ & $0.6891$ & $0.0030$ \\
Qwen3-8B / MATH-500 & $0.7416$ & $0.7416$ & $0.7416$ & $0.7416$ & $0.7416$ & $0.7416$ & $0.7416$ & $0.7416$ & $0.7420$ & $0.6718$ & $0.0004$ \\
SmolLM3-3B / MATH-500 & $0.7173$ & $0.7173$ & $0.7173$ & $0.7173$ & $0.7170$ & $0.7135$ & $0.5226$ & $0.4437$ & $0.6319$ & $0.6332$ & $0.0003$ \\
gpt-oss-20b / MATH-500 & $0.5983$ & $0.5983$ & $0.5983$ & $0.5983$ & $0.5983$ & $0.5983$ & $0.5983$ & $0.5983$ & $0.5983$ & $0.5537$ & $0.0000$ \\
Qwen3-1.7B / GPQA-Diamond & $0.6028$ & $0.6028$ & $0.6028$ & $0.6028$ & $0.6028$ & $0.6028$ & $0.6028$ & $0.6028$ & $0.6028$ & $0.5618$ & $0.0000$ \\
Qwen3-4B / GPQA-Diamond & $0.7027$ & $0.7027$ & $0.7027$ & $0.7027$ & $0.7027$ & $0.7027$ & $0.7027$ & $0.7026$ & $0.7077$ & $0.6402$ & $0.0049$ \\
Qwen3-8B / GPQA-Diamond & $0.7179$ & $0.7179$ & $0.7179$ & $0.7179$ & $0.7179$ & $0.7179$ & $0.7179$ & $0.7179$ & $0.7198$ & $0.6345$ & $0.0020$ \\
SmolLM3-3B / GPQA-Diamond & $0.5577$ & $0.5577$ & $0.5577$ & $0.5577$ & $0.5578$ & $0.5575$ & $0.5230$ & $0.5071$ & $0.5059$ & $0.5047$ & $0.0000$ \\
gpt-oss-20b / GPQA-Diamond & $0.6017$ & $0.6017$ & $0.6017$ & $0.6017$ & $0.6017$ & $0.6017$ & $0.6017$ & $0.6017$ & $0.6017$ & $0.4197$ & $0.0000$ \\
\bottomrule
\end{tabular}
}
\vspace{0.2em}
\caption{Temperature sweep for AUROC. The endpoint $\tau\to 0$ corresponds to hard layer selection, whereas $\tau\to\infty$ corresponds to uniform layer weighting.}
\label{tab:tau_sweep}
\end{table*}

\section{Related Work}
\label{app:related_work}

\paragraph{Test-time computation and surface-trace measures.}
A broad line of work improves reasoning by increasing or allocating inference-time computation, through CoT prompting, test-time scaling, budget forcing, or reasoning-oriented reinforcement learning \citep{wei2023chainofthoughtpromptingelicitsreasoning, snell2025scaling, muennighoff2025s1, Guo_2025, yang2025qwen3technicalreport}. 
This line primarily asks how to elicit, allocate, or exploit additional reasoning computation. 
A separate line asks whether the resulting trace is actually useful. 
Longer traces are not always better, since models may overthink, and accuracy can degrade when reasoning chains become excessively long or noisy \citep{zhang2025reasoningmodelsknowtheyre, eisenstadt2025overclockingllmreasoningmonitoring, wu2025lessunderstandingchainofthoughtlength}. 
Recent internal-effort measures therefore move beyond token counts, for example, deep-thinking tokens identify positions whose predictions are substantially revised in deeper layers \citep{chen2026thinkdeepjustlong}. 
\stalt{} follows this diagnostic direction rather than the inference-scaling direction. 
It does not prescribe a decoding policy or rely on the length of the visible CoT.
Instead, it measures whether a generated trajectory exhibits large local transitions in the model's hidden states.

\paragraph{Output-free verification and latent trajectory summaries.}
Most reliability methods estimate answer quality from observable or externally supplied signals, such as verbalized confidence, token probabilities, entropy, semantic uncertainty, self-consistency, or process reward models \citep{lin2022teachingmodelsexpressuncertainty, tian-etal-2023-just, kuhn2023semantic, xiong2024llmsexpressuncertaintyempirical, lightman2023let}. 
Hidden-state methods are closer to our goal because they inspect the computation that produces the answer. 
Early work showed that truthfulness can be decoded from internal activations \citep{azaria-mitchell-2023-internal, liu-etal-2023-cognitive}, and recent methods use hidden states for self-verification or efficient test-time control. 
CoE treats progressive hidden states as a latent thinking path for output-free self-evaluation \citep{wang2025latent}; CLUE clusters hidden-state deltas from past successful and failed traces to build a non-parametric verifier \citep{liang2025cluenonparametricverificationexperience}; hidden-state probes and internal-circuit methods train lightweight predictors for intermediate-answer correctness or failure detection \citep{zhang2025reasoningmodelsknowtheyre, ghasemabadi2026llmspredictfailuresselfawareness}; and STEP uses a learned step scorer to prune unpromising traces during parallel test-time scaling \citep{liang2026hiddenstatesearlysignals}. 
Very recent latent-trajectory approaches are especially close, since they also compute scalar signals from hidden-state motion during reasoning \citep{vilas2025tracingtraceslatenttemporal, bi2025cotkineticstheoreticalmodelingassessing}. 
The difference is primarily in granularity and aggregation.
These methods reduce hidden-state motion to trajectory-level summaries, such as start-to-end displacement, accumulated change, alignment, or kinetic-energy scores, for verification or selection. 
In contrast, \stalt{} is a training-free statistic that preserves token-local temporal variation and uses layer-local change to emphasize where the transition is concentrated.

\paragraph{Decoupling Knowledge and Active Computation.}
Differentiating genuine reasoning from passive memory retrieval is a growing challenge.
Analyses using cognitive dual-system theory suggest that LLM generation decouples into distinct spatial phases: static knowledge retrieval is largely localized in lower layers, while active reasoning dynamically engages higher layers \citep{yang2026decoupling}.
Furthermore, \citet{wu2025knowledgereasoningcloselook} showed that supervised fine-tuning often enhances factual recall at the expense of active, uncertainty-reducing reasoning paths.
Our observations are consistent with this spatial asymmetry. 
The layer-weighting mechanism in \stalt{} gives more weight to depths where within-token representational changes are large, while preserving the temporal movement across decoding steps.

\section{Implementation Details}
\label{app:experimental_setup}
\subsection{Models}
We use the official Hugging Face implementations for all models. 
All inference is performed in bfloat16 precision without weight quantization.

\subsection{Dataset}
We evaluate on five benchmarks chosen to cover arithmetic word problems, competition-style mathematics, expert-level science question answering, broad professional knowledge, and controlled numeric reasoning. 
We use the official Hugging Face implementations of all datasets.

\textbf{GSM8K} \citep{cobbe2021gsm8k} contains linguistically diverse grade-school math word problems that typically require a short sequence of elementary arithmetic operations to reach a single numeric answer. The problems are simple at the level of required mathematical concepts but still require multi-step decomposition. We use the 1,319-example test split.

\textbf{MATH-500} is a curated 500-problem subset of the MATH benchmark \citep{math500}. 
The problems are substantially harder than GSM8K and span competition-style secondary-school mathematics, including algebra, geometry, number theory, counting, and precalculus.
These items usually require longer derivations, symbolic manipulation, and nontrivial intermediate lemmas before producing a final boxed answer, making the benchmark useful for probing sustained mathematical reasoning. 
We use the 500-example test split.

\textbf{GPQA-Diamond} \citep{rein2024gpqa} is a multiple-choice benchmark of very hard graduate-level questions in biology, physics, and chemistry written and validated by domain experts. 
The Diamond split is designed to be especially challenging and resistant to shallow retrieval or elimination strategies, so success requires extended scientific reasoning over specialized knowledge rather than routine factual recall. 
We use the standard 198-question Diamond split, comprising 93 chemistry, 86 physics, and 19 biology questions.

\textbf{MMLU-Pro} \citep{NEURIPS2024_ad236edc} is a more difficult and more reasoning-oriented version of MMLU \citep{hendryckstest2021}, with more answer options and substantially stronger distractors. 
It spans 14 disciplines and mixes problems that demand explicit reasoning with problems that are closer to professional knowledge retrieval, which makes it suitable for testing whether the hidden-state signature weakens outside clearly reasoning-intensive settings. 
We use the 12,032-example test set, with subject counts of 1,351 in math, 1,299 in physics, 1,132 in chemistry, 1,101 in law, 969 in engineering, 924 in other, 844 in economics, 818 in health, 798 in psychology, 789 in business, 717 in biology, 499 in philosophy, 410 in computer science, and 381 in history.

\textbf{s1K-1.1} \citep{muennighoff2025s1} is a 1,000-example reasoning set accompanied by long model-generated traces. 
The questions are drawn from math-focused sources, and for our intervention experiments we retain the 472 integer problems because they admit exact numeric verification without ambiguity in answer matching. 
This benchmark therefore provides a controlled setting in which changes in hidden-state dynamics can be compared against a reliable correctness signal.

\subsection{Inference Configuration}

\textbf{Prompts} \\
The specific base prompts used for each benchmark are detailed below. 
We replace \texttt{\{question\}} with the actual problem statement and \texttt{\{subject\}} with the relevant domain category.

\vspace{1em}

\begin{promptbox}{MATH-500 and GSM8K}
Solve this math problem. Give the reasoning steps before giving the final answer inside \textbackslash boxed\{X\}. \\
Q: \{question\}
\end{promptbox}

\begin{promptbox}{GPQA-Diamond}
The following are multiple choice questions. Think step by step and then finish your answer with ``the answer is (X)'' where X is the correct letter choice. \\
Q: \{question\}
\end{promptbox}

\begin{promptbox}{MMLU-Pro}
The following are multiple choice questions about \{subject\}. Think step by step and then finish your answer with ``the answer is (X)'' where X is the correct letter choice. \\
Q: \{question\}
\end{promptbox}

\begin{promptbox}{s1K-1.1}
Solve this math problem. Reason step by step, and write your final answer as an integer inside \textbackslash boxed\{\}. \\
Q: \{question\}
\end{promptbox}

\vspace{1em}
\textbf{In-Context CoT Scaffolding} \\
For evaluations of CoT scaffolding, we append the following exemplar format to the base prompt, where \texttt{\{cot\}} contains the step-by-step reasoning and final answer depending on the setting:

\begin{promptbox}{Oracle CoT Exemplar}
You can refer to example reasoning and answer. \\
Example: \\
\{cot\}
\end{promptbox}

\textbf{Decoding Parameters} \\
Unless otherwise noted, generation uses a temperature of 0.7, top-$p$ of 0.95, and a maximum generation length of 16384 tokens.
For Qwen2.5-Math-7B alone, we reduce the maximum generation length to 4096 tokens because of its context-length constraint.

\subsection{Baselines}

We evaluate \stalt{} against length-based, output-space, embedding-space, and
unweighted hidden-state baselines. For a generated response with token length
$T$, let $c_t$ denote the token generated at step $t$, and let $p_t(v)$ be the
model's next-token probability for vocabulary item $v \in V$, computed from the
full-vocabulary logits recorded at that step. Unless otherwise stated, all
baseline scores are oriented so that larger values indicate a higher predicted
likelihood of correctness.

\paragraph{Number of Generated Tokens}
We use the negative response length as a simple length-based predictor:
\[
s_{\mathrm{len}} = -T.
\]
This baseline tests whether shorter reasoning traces are more likely to be
correct in our evaluation setting.

\paragraph{Maximum Token Probability}
We compute the average maximum next-token probability:
\[
s_{\mathrm{maxprob}}
=
\frac{1}{T}\sum_{t=1}^{T}\max_{v \in V} p_t(v).
\]
This score measures how concentrated the model's token-level predictive
distribution is during generation.

\paragraph{Perplexity}
We use negative sequence perplexity as the predictive score:
\[
s_{\mathrm{ppl}}
=
-\exp\left(
-\frac{1}{T}\sum_{t=1}^{T}\log p_t(c_t)
\right).
\]
The negative sign orients the score so that larger values correspond to lower
perplexity and therefore higher token-level confidence.

\paragraph{Entropy}
We compute the negative average token-level entropy:
\[
s_{\mathrm{ent}}
=
-\frac{1}{T}\sum_{t=1}^{T}
\sum_{v \in V} -p_t(v)\log p_t(v).
\]
This score is larger when the model's predictive distributions are sharper.

\paragraph{Chain-of-Embedding}
We include CoE-R and CoE-C from \citet{wang2025latent} as embedding-space
verification baselines. These methods summarize a generation using the
layer-wise hidden-state chain
\[
H = h_0 \rightarrow h_1 \rightarrow \cdots \rightarrow h_L,
\]
where $h_l$ denotes the token-averaged hidden state at layer $l$. Let
\[
M(h_l,h_{l+1})=\lVert h_{l+1}-h_l\rVert_2,
\qquad
A(h_l,h_{l+1})=
\arccos
\left(
\frac{h_{l+1}^{\top}h_l}
{\lVert h_{l+1}\rVert_2 \lVert h_l\rVert_2}
\right).
\]
CoE-R combines normalized magnitude and angle changes as
\[
s_{\mathrm{CoE\mbox{-}R}}
=
\frac{1}{L}
\sum_{l=0}^{L-1}
\left(
\frac{M(h_l,h_{l+1})}{M(h_0,h_L)}
-
\frac{A(h_l,h_{l+1})}{A(h_0,h_L)}
\right).
\]
CoE-C aggregates the same adjacent-layer transitions in the complex plane:
\[
s_{\mathrm{CoE\mbox{-}C}}
=
\frac{1}{L}
\sqrt{
\left(
\sum_{l=0}^{L-1} M(h_l,h_{l+1})\cos A(h_l,h_{l+1})
\right)^2
+
\left(
\sum_{l=0}^{L-1} M(h_l,h_{l+1})\sin A(h_l,h_{l+1})
\right)^2
}.
\]
We use these scores as label-free hidden-state baselines without fitting any
additional classifier.

\paragraph{Unweighted Hidden-State Amplitudes}
We include two direct ablations of the components used by \stalt{}.
The first averages temporal hidden-state changes uniformly across all layers:
\[
s_{\Delta_{\mathrm{time}}}
=
\frac{1}{(T-1)(L+1)}
\sum_{t=2}^{T}\sum_{l=0}^{L}
\Delta_{\mathrm{time}}[t,l].
\]
The second averages layer-wise hidden-state changes uniformly across all token
positions:
\[
s_{\Delta_{\mathrm{layer}}}
=
\frac{1}{TL}
\sum_{t=1}^{T}\sum_{l=1}^{L}
\Delta_{\mathrm{layer}}[t,l].
\]
These baselines test whether the layer-weighted temporal integration in
\stalt{} improves over simple global averages of the underlying hidden-state
transition magnitudes.
\subsection{Supervised Fine-Tuning Configuration}
\label{app:sft_config}

For the supervised fine-tuning intervention, we train Qwen3-4B on the DeepSeek-R1 traces and responses for the integer problems in s1K-1.1. 
The training hyperparameters are summarized in Table~\ref{tab:sft_hparams}.

\begin{table}[t]
\centering
\small
\caption{Hyperparameters used for supervised fine-tuning.}
\label{tab:sft_hparams}
\begin{tabular}{ll}
\toprule
\textbf{Hyperparameter} & \textbf{Value} \\
\midrule
Training epochs & 10 \\
Learning rate & $5.0\times10^{-7}$ \\
Learning-rate schedule & Cosine \\
Optimizer & AdamW \\
Adam betas & $(0.9, 0.95)$ \\
Weight decay & 0.01 \\
Warmup ratio & 0.05 \\
Maximum gradient norm & 1.0 \\
Maximum sequence length & 16,384 tokens \\
Batch size & 64 \\
Precision & bfloat16 \\
\bottomrule
\end{tabular}
\end{table}

\subsection{Compute Resources}
\label{app:computing_resources}
Model inference and training were performed on a single NVIDIA H200 GPU with 141 GB of memory. 
The only exception was hidden-state extraction for gpt-oss-20b, which required two NVIDIA H200 141 GB GPUs. 
We generated output traces with vLLM \citep{kwon2023efficient} and used Hugging Face Transformers \citep{wolf-etal-2020-transformers} to extract hidden states from the generated traces and compute the proposed and baseline metrics. 
The supervised fine-tuning was implemented using TRL \citep{vonwerra2020trl}.

\section{Declaration of LLM Usage}
\label{app:llm_usage}

This paper studies large language models and large reasoning models as experimental subjects. 
The evaluated models, prompts, decoding settings, and hidden-state extraction procedures are described in Sections~\ref{sec:observations}--\ref{sec:perturbation} and Appendix~\ref{app:experimental_setup}.

The authors used LLM-based tools for manuscript proofreading and assistance in drafting experimental code. 
All experimental code, outputs, figures, references, and scientific claims were reviewed, verified, and finalized by the authors.

\section{Broader Impacts}
\label{sec:broader_impacts}

This work characterizes hidden-state trajectory patterns associated with successful reasoning in large reasoning models.
By making aspects of latent computation measurable, the analysis may support failure diagnosis and comparison of reasoning behavior beyond final-answer accuracy or trace length.
A possible risk is that internal trajectory statistics are treated as evidence of reliability, or optimized as objectives, without independent validation of task-grounded reasoning.
We therefore regard \stalt{} as a diagnostic for model analysis rather than a deployment criterion; uses in self-evaluation or reward construction should be validated against external task- and domain-specific criteria.



\end{document}